\documentclass{article}

\usepackage{natbib}

\usepackage{breakcites}
\usepackage[margin=0.8in]{geometry}

\usepackage{authblk}
\usepackage{algorithm}
\usepackage{algorithmic}

\usepackage{graphicx}
\usepackage{amsmath}
\usepackage{bm}
\usepackage{caption}
\usepackage{subcaption}
\usepackage{wrapfig}

\usepackage[utf8]{inputenc} % allow utf-8 input
\usepackage[T1]{fontenc}    % use 8-bit T1 fonts
\usepackage[x11names]{xcolor}
\usepackage{hyperref}       % hyperlinks
\hypersetup{
    colorlinks=true,
    linkcolor=blue,
    citecolor=Green3
}
\usepackage{url}            % simple URL typesetting
\usepackage{booktabs}       % professional-quality tables
\usepackage{amsfonts}       % blackboard math symbols
\usepackage{nicefrac}       % compact symbols for 1/2, etc.
\usepackage{microtype}      % microtypography
\usepackage{xcolor}         % colors

\def\keywordname{{\bfseries \emph{Keywords}}}%
\def\keywords#1{\par\addvspace\medskipamount{\rightskip=0pt plus1cm
\def\and{\ifhmode\unskip\nobreak\fi\ $\cdot$
}\noindent\keywordname\enspace\ignorespaces#1\par}}

\title{Active Sensing with Predictive Coding and Uncertainty Minimization}

\author[1, 2, 3]{\textbf{Abdelrahman Sharafeldin}}
\author[1, 2]{\textbf{Nabil Imam}}
\author[1, 3]{\textbf{Hannah Choi}}
\affil[1]{ML@GT, Georgia Institute of Technology}
\affil[2]{School of Computational Science and Engineering, Georgia Institute of Technology}
\affil[3]{School of Mathematics, Georgia Institute of Technology}
\affil[ ]{\texttt {\{abdo.sharaf,nimam6,hannahch\}@gatech.edu}}

\date{}

\begin{document}

\renewcommand{\abstractname}{Summary}

\maketitle

\begin{abstract}
  We present an end-to-end procedure for embodied exploration inspired by two biological computations: predictive coding and uncertainty minimization. The procedure can be applied to exploration settings in a task-independent and intrinsically driven manner. We first demonstrate our approach in a maze navigation task and show that it can discover the underlying transition distributions and spatial features of the environment. Second, we apply our model to a more complex active vision task, where an agent actively samples its visual environment to gather information. We show that our model builds unsupervised representations through exploration that allow it to efficiently categorize visual scenes. We further show that using these representations for downstream classification leads to superior data efficiency and learning speed compared to other baselines while maintaining lower parameter complexity.  Finally, the modularity of our model allows us to probe its internal mechanisms and analyze the interaction between perception and action during exploration.
\end{abstract}

\keywords{Predictive coding \and active vision \and embodied exploration \and generative model \and variational inference}

\section{Introduction}
\label{introduction}
Biological organisms interact with the world in cycles of perception and action. These two processes are intertwined and interact with one another to guide animal behavior \citep{guillery_1, guillery_2, active_inference_embodied, FristonTheFP}. Visual perception, for example, is not passive. Rather, we actively sample our visual field in search of information, a process called active vision in neuroscience and psychology \citep{yarbus, hayhoe_eye_movements, ian_visual_fix, land_looking_acting, yang_bas, Friston_perceptions}. Similarly, an animal navigating a maze explores its environment and builds accurate representations of its structure, subsequently using them for various goal-directed tasks. In contrast, most models of artificial intelligence (AI) treat perception and action as separate processes and aim to optimize performance with respect to task-specific objectives. For example, visual recognition in machine learning often utilizes convolutional neural networks (CNNs) \citep{alex_net, resnet_paper}, which passively receive entire images as input, to directly maximize classification accuracy on a given dataset. Another example is reinforcement learning \citep{Sutton1998}, where actions are chosen primarily to maximize extrinsic reward, without accounting for the agent's intrinsic motivations and priors. By leveraging insights from neuroscientific theories of perception and action, we can develop embodied AI models that actively explore their environment and interact with the physical world \citep{zador_catalyzing}.

In this work, we integrate two theories from systems neuroscience to develop a combined perception-action model for intrinsically driven active sensing. We base the perception component of our model on the theory of predictive coding \citep{rao_ballard}. According to predictive coding, the brain maintains a generative model of the world \citep{bruno_ref1} which it uses to predict its sensory input. The goal of perception, therefore, is to infer the latent states of this generative model \citep{bruno_ref2} so as to minimize prediction error. The action component of our model is based on the proposition that the brain minimizes uncertainty of inferred latent states during exploratory behavior \citep{butko_infomax, butko_ipomdp, little_sommer, FristonTheFP, rafa_ref1}. Due to the intractability of the uncertainty reduction objective (or equivalently, the information gain objective), most models that optimize it rely either on sample-inefficient reinforcement learning methods or on restrictive assumptions that make it easier to evaluate. In our approach, we use a deep generative model based on predictive coding that allows us to optimize a Monte Carlo (MC) approximation to the information gain objective in a fully differentiable manner without assuming explicit knowledge of the true generative model of the environment. We show that this approximation, even when done in a greedy fashion, leads to a highly efficient exploration strategy. 

Our model integrates perception and action within an end-to-end differentiable procedure and can be generally applied to any exploration setting in a task-independent manner without the need for extrinsic reward signals. To illustrate, we evaluate our model on two sensorimotor tasks. First, we test the model on a simple maze navigation task with noisy transitions and show that it explores the environment more efficiently than both random exploration and visitation-count-based Boltzmann exploration. We show that our agent learns an exploration policy that enables its perception model to quickly discover the underlying transition distributions of the maze. Second, we apply our model to the more complicated task of active vision. In this task, the model has a band-limited sensor which it uses to perceive small patches of a hidden image through a limited number of fixations. We show that, despite its band-limited perception, the model is able to learn the spatial relationships between pixels of a given image, as demonstrated by its ability to generate full meaningful images by combining smaller generated patches at different locations. Furthermore, we show that, although these representations are learned unsupervised, they enable a downstream classifier to quickly reach high test performance with fewer training data and lower parameter complexity. We compare these results to a feedforward network receiving full images as well as to other popular baselines from RL literature including the Recurrent Attention Model (RAM) \citep{mnih_ram}, VIME \citep{vime}, and Plan2Explore \citep{p2e}. 

\begin{figure}[h]
    \centering
    \begin{subfigure}[c]{0.4\textwidth}
        \includegraphics[width=\textwidth]{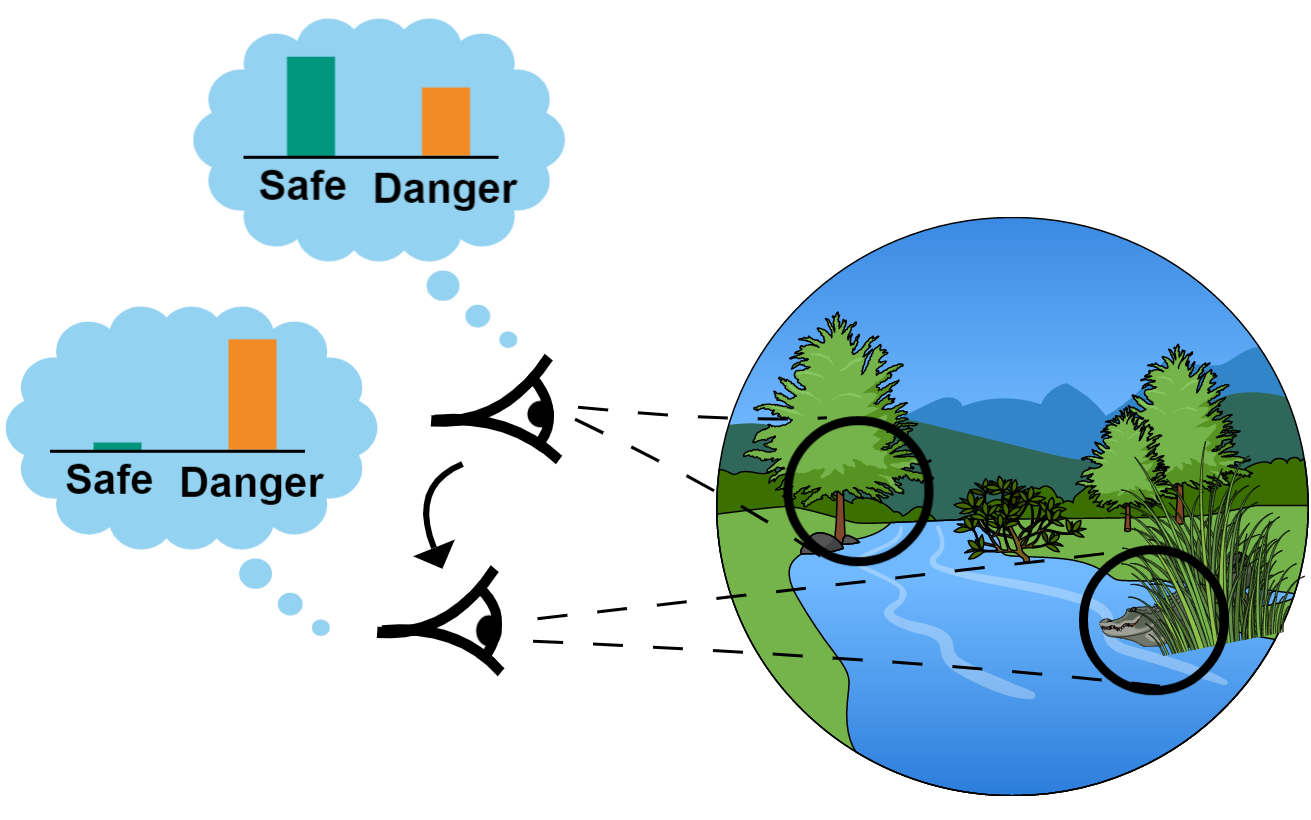}
        \caption{}
        \label{idea_demo_av}
    \end{subfigure}
    \;\;\;\;\;\;\;\;\;\;
    \begin{subfigure}[c]{0.45\textwidth}
        \includegraphics[width=\textwidth]{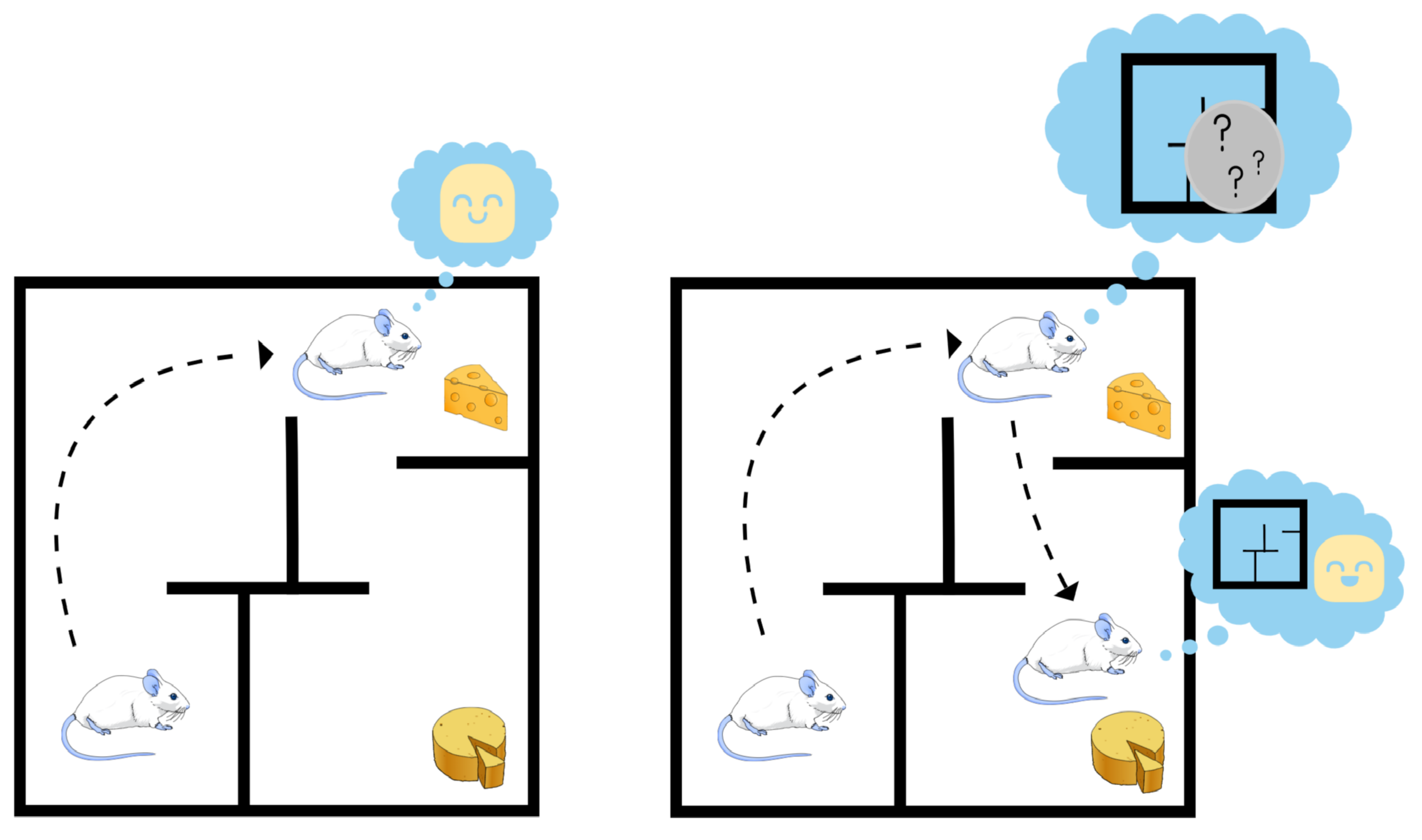}
        \caption{}
        \label{fig:idea_demo_1}
    \end{subfigure}
    \caption{Traditional versus biological models of perception and action. (a) we actively sample visual scenes to infer hidden states, in contrast to standard ML models which assume passive perception. (b) biological systems have an intrinsic drive to actively explore the environment and build internal models of it; in contrast, traditional RL models are primarily guided by extrinsic reward.}
    \label{fig:idea_demo} 
\end{figure}

Our model selects actions that are purely intrinsically driven to minimize its uncertainty about the environment (Figure \ref{fig:idea_demo}). In that sense, the action component is blind to the task at hand and only has access to the perception model's internal states. This makes its ability to perform well on image classification quite remarkable. Importantly, the quality of the perceptual representations depends on the action selection strategy, highlighting the perception-action relationship captured by our framework. We quantify this effect by measuring the mutual information between image categories and learned representations under different action strategies. Furthermore, the modular structure of our model facilitates interpretability, allowing us to probe its mechanisms and representations during exploration and providing us with insights into the possible neural computations utilized in biological systems. For example, we show that during active vision, the model learns representations that reflect the properties of the data and the structure of the task. 
Our approach demonstrates the promise of integrating neuroscientific theories of perception and action into embodied AI agents, and we hope that it will motivate more research in this area. A survey of related work is provided in Appendix \ref{appendix:related_work}. 

\section{Model Description}
\label{model_desc}
\subsection{General approach for active exploration}
Our approach consists of two components: a perception model, which is based on predictive coding \citep{rao_ballard}, and an action model, which selects actions that reduce the perception model's uncertainty about inferred states. For perception, we rely on a generative model of the world that can be learned through experience. Perceiving an observation corresponds to inverting this model to infer the hidden states of the world that gave rise to that observation. Therefore, the first step when applying our approach to a given problem is to specify a reasonable generative model; in biological systems, this corresponds to niche-specific priors. The second step is to specify a method for learning and inference in this model. To this end, we use variational autoencoders (VAEs) \citep{kingma_vae} to perform amortized variational inference on a generative model parameterized by neural networks. We describe the connection between VAEs and predictive coding in Appendix \ref{appendix:pc_vae}. In essence, given a simple generative model in which a latent variable $z$ gives rise to a set of observed variables $x_{\leq t}$ at times up to time $t$, the goal of variational inference \citep{jordan_variational_methods} is to find an \textit{approximate} posterior $q(z|x_{\leq t})$ which maximizes the objective 
\begin{equation}\label{eqn:elbo}
    \mathbb{E}_{z\sim q(\cdot|x_{\leq t})}[\log{p(x_{\leq t}|z)}] - D_{KL}(q(z|x_{\leq t})||p(z))
\end{equation}
where $D_{KL}$ is the KL divergence. This objective is the evidence lower bound or ELBO. Amortized learning can be done by using neural networks to parameterize the distributions in \ref{eqn:elbo} and optimizing the objective with gradient descent. Inference in this case corresponds to a simple forward pass through a neural network. 

The second component of our approach is the action model, which relies on uncertainty reduction measured using Shannon entropy. As such, an application of our approach requires a method of computing the entropy of the posterior distribution $q(z|x_{\leq t})$ inferred by the perception model. Actions are then selected to maximize the reduction in uncertainty as represented by the following score function 
\begin{equation}\label{eqn:uncertainty_red}
    \text{Score}(a) = H(q(z|x_{\leq t})) - \mathbb{E}_{x_{t+1} \sim p(\cdot|a, x_{\leq t})}\left[ H(q(z|x_{t+1}, x_{\leq t}))\right]
\end{equation}
where $H$ denotes the Shannon entropy. The first term in \ref{eqn:uncertainty_red} represents the agent's current uncertainty about $z$. The second term represents the \textit{expected} uncertainty if action $a$ is executed. The expectation in the second term is taken with respect to $\textit{fictitious}$ future states drawn from the agent's current transition distribution $p(x_{t+1}|x_{\leq t}, a)$. In discrete action and state spaces, this score function can be evaluated directly for each action, but it becomes intractable in continuous state and action spaces. To work around this issue, we use a Monte Carlo (MC) approximation to the expectation in the second term and take advantage of a deep generative model to compute the score in a differentiable manner. We refer to this action selection strategy as Bayesian Action Selection (BAS). 

\subsection{Exploration in controllable Markov chains}
As a proof of concept, we first demonstrate our model in the setting of discrete state and action spaces. Specifically, we develop an instance of the general framework described above for Controllable Markov Chains (CMC) \citep{gimbert_cmcs, little_sommer}. A CMC is essentially a Markov decision process (MDP) but without the specification of a reward function. It is formally defined as a 3-tuple $(\mathcal{S}, \mathcal{A}, \mathcal{P})$, where: $\mathcal{S}$ is a set of finite states, e.g. the set of possible locations in a maze; $\mathcal{A}$ is a finite set of allowable actions, e.g. movement directions; $\mathcal{P}$ is a 3-dimensional kernel of transition probabilities $p: \mathcal{S} \times \mathcal{A} \rightarrow \mathcal{D}(\mathcal{S})$, where $\mathcal{D}(\mathcal{S})$ is the set of probability distributions on $\mathcal{S}$. That is, $\mathcal{P}$ is a $|\mathcal{S}| \times |\mathcal{A}| \times |\mathcal{S}|$ matrix containing the probabilities $\mathcal{P}_{s, a, s'} = p(s'|s, a)$, where $s$ is the current state, $a$ is the action taken, and $s'$ is the resulting next state. 

The goal of an agent in this setting is to efficiently explore the environment and learn an estimate, $\hat{\mathcal{P}}$, of the underlying transition probability matrix $\mathcal{P}$. This setting models the embodiment of the agent because, at any given time, the agent's interaction with the world is restricted by its current state. The specific CMC in which we test our model is a maze environment similar to that used in \cite{little_sommer}. In a given environment, there are $N = n^2$ states corresponding to locations in an $n \times n$ maze, and 4 actions corresponding to the cardinal directions (up, down, right, and left). Each action produces a noisy translation, with more bias towards to the cardinal direction associated with that action. All transitions that do not correspond to a one-step translation (i.e., a neighboring state) are assigned a probability of zero. The mazes are randomly generated and the probability distributions in $\mathcal{P}$ are drawn from a Dirichlet distribution with concentration parameters $\alpha = 0.25$ for states with non-zero probability. 

\subsubsection{Perception}
\begin{figure}
    \centering
    \begin{subfigure}[c]{0.45\textwidth}
        \includegraphics[width=\textwidth]{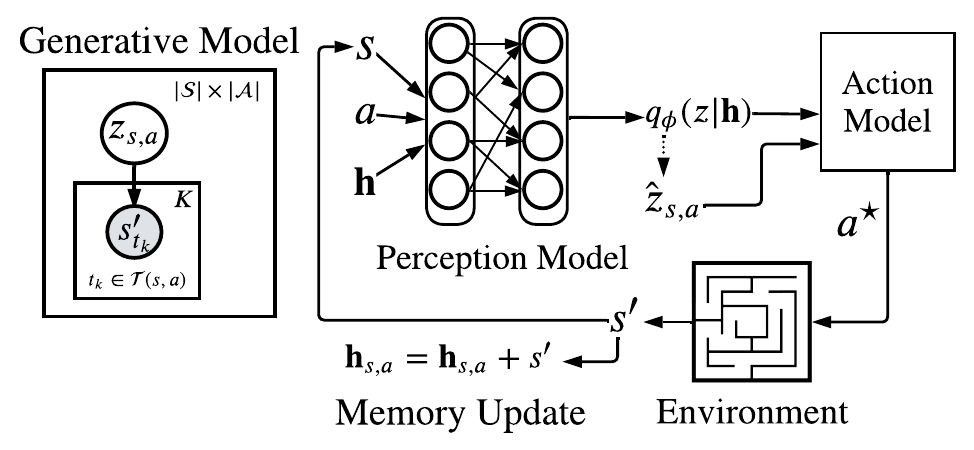}
        \caption{}
        \label{fig:cmc_model}
    \end{subfigure}
    \;
    \begin{subfigure}[c]{0.5\textwidth}
        \includegraphics[width=\textwidth]{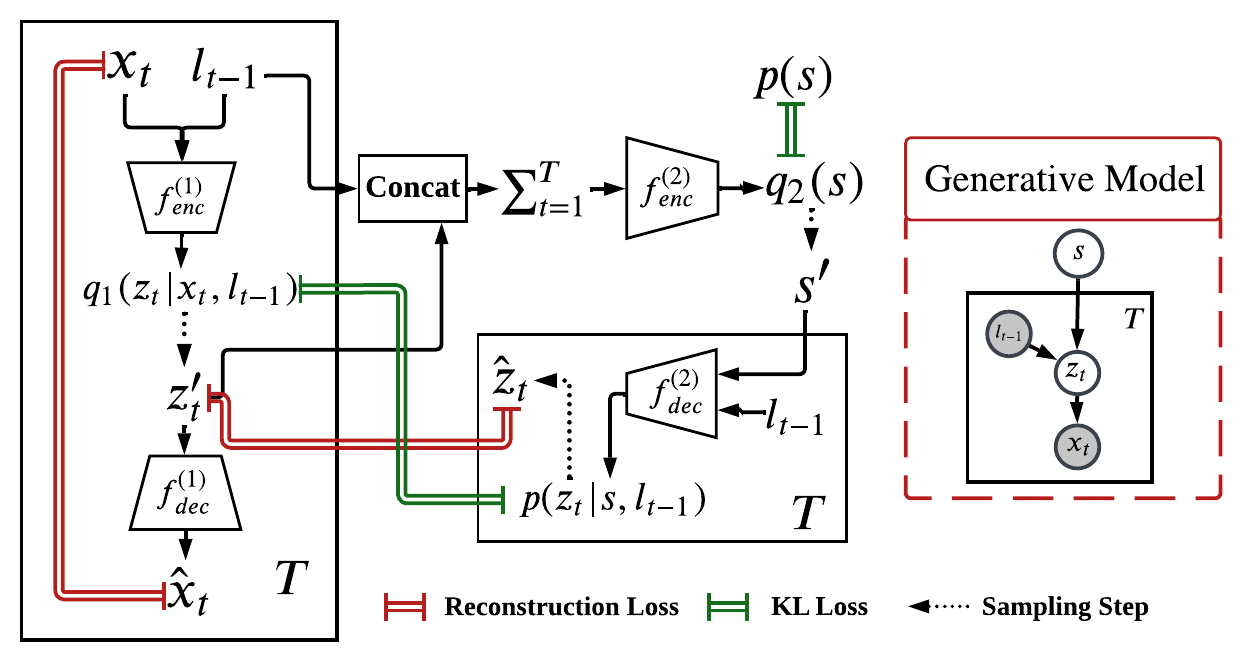}
        \caption{}
        \label{fig:av_model}
    \end{subfigure}
    \caption{Generative models and architectures for the active exploration agents in (a) CMCs, and (b) active vision. Shaded and unshaded circles represent observed and latent variables, respectively. In (b), $f^{(1)}_{\{enc,\;dec\}}$ refer to the encoder and decoder networks of the lower-level VAE, while $f^{(2)}_{\{enc,\;dec\}}$ refer to those of the higher-level VAE. Plate notation is used for the parts that are repeated for every time step $t$ up to the total number of allowed fixations $T$.}
\end{figure}

We begin by specifying a generative model for this task. This generative model constitutes the perception component and is learned from observations collected by the agent. Whenever the agent visits a state $s$ and takes action $a$, the observation consists of the resulting state $s'$. The agent's goal is to infer the distribution $\hat{p}(:|s, a)$ for each state-action combination that best explains all observations collected when that combination was visited. Let $z_{s,a} = \hat{p}(:|s, a)$ denote the model's estimated distribution of the next state after executing action $a$ in state $s$. Then, the generative model contains $|\mathcal{S}| \times |\mathcal{A}|$ latent variables, $\{z_{s, a}\}_{s\in\mathcal{S}, a\in\mathcal{A}}$, each corresponding to a state-action combination. If we let $\mathcal{T}(s, a)$ be the set of times at which state $s$ was visited and action $a$ was taken, then each latent variable $z_{s, a}$ gives rise to $K = |\mathcal{T}(s, a)|$ observations (or next states), denoted by $\{s'_{t_k}\}_{t_k \in \mathcal{T}(s, a)}$. This is illustrated in Figure \ref{fig:cmc_model}.

Let $\mathcal{H} = \{(s_t, a_t, s'_t)\}_{t=1}^{T}$ denote the full history of experiences collected by the agent, and $\mathcal{H}_{s, a} = \{(s, a, s'_{t})\}_{t\in\mathcal{T}(s, a)}$ be the subset of $\mathcal{H}$ containing experiences in state $s$ when action $a$ was taken. Because the latent variables are independent, the variational posterior in this model can be expressed as 
\begin{equation}\label{eqn:cmc_posterior}
    q\left(\{z_{s, a}\}\right|\mathcal{H}) = \prod_{s\in \mathcal{S}, a\in\mathcal{A}} q\left(z_{s, a}|\mathcal{H}_{s, a}\right)
\end{equation}

To simplify notation, we will drop the subscript $(s, a)$, but it should be clear that, in what follows, the latent variables $z$ and the histories $\mathcal{H}$ depend on the specific state-action combination. For a single latent variable $z$, the corresponding ELBO can be expressed as 
\begin{align}
    \log{p(\mathcal{H})} \geq&\; \mathbb{E}_{z \sim q(\cdot|\mathcal{H})}\left [ \log p(z, \mathcal{H}) - \log{q(z|\mathcal{H})} \right]\\
    =&\; \mathbb{E}_{z\sim q(\cdot|\mathcal{H})}\left[ \log{p(\mathcal{H}|z)}\right] - D_{KL}\Big(q(z|\mathcal{H})||p(z)\Big)\label{eqn:cmc_single_elbo}
\end{align}

 The first term in \ref{eqn:cmc_single_elbo} corresponds to the likelihood of the observations given the inferred transition distribution $z$, while the second term controls the deviation of that distribution from the model's own prior $p(z)$. We assume the prior for all latent variables is a Dirichlet distribution with concentration parameters $\alpha = 1$ for all states in $\mathcal{S}$. The posterior $q(z|\mathcal{H})$ is also assumed to be a Dirichlet distribution with a concentration parameter $\bm{\alpha}$ that is the output of a simple feedforward neural network with parameters $\phi$. 

To compute the quantities in \ref{eqn:cmc_single_elbo}, we need a representation of $s$, $a$, and $\mathcal{H}$ that facilitates updating the agent's history with new observations. To achieve this, we use one-hot vectors to represent the states and actions. We represent the full history of the agent as a $|\mathcal{S}| \times |\mathcal{A}| \times |\mathcal{S}|$ matrix whose entries represent visit counts to state $s'$ from state $s$ when action $a$ was taken. That is, we represent the history $\mathcal{H}_{s, a}$ as the vector $\bm{h}_{s, a} \in \mathbb{N}_{0}^{|\mathcal{S}|}$ with entries
\begin{equation}
    \bm{h}_{s, a}(i) = \sum_{t\in\mathcal{T}(s, a)} \delta(s'_t, \mathcal{S}(i))\;\;\;\; \text{for} \; i = 1, 2, ..., |\mathcal{S}|
\end{equation}

This representation makes updating the history given new observations a simple summation operation. That is, when the agent takes action $a$ in state $s$ and receives new observation $s'$, the history can simply be updated as $\bm{h}_{s, a} \leftarrow \bm{h}_{s, a} + s'$. Now, the log-likelihood in \ref{eqn:cmc_single_elbo} can be computed as
\begin{equation}
    \log p(\mathcal{H}|z) = \log \prod_{i = 1}^{|\mathcal{S}|} z(i)^{\bm{h}_{s, a}(i)} = \sum_{i=1}^{|\mathcal{S}|}\bm{h}_{s, a}(i)\log{z(i)}
\end{equation}
Amortized Inference in this model is performed by the perception network $\phi$ which receives as input the current state $s$, action $a$, and history vector $\bm{h}_{s, a}$ and outputs a concentration parameter $\bm{\alpha}$ that parameterizes the Dirichlet posterior distribution $q_{\phi}(z|\mathcal{H})$. The inferred distribution $z$ can be obtained by drawing a reparameterized sample from $q_{\phi}(z|\mathcal{H})$ \cite{kingma_vae}. Note that the full ELBO for this model should be the sum of the terms in \ref{eqn:cmc_single_elbo} over all state-action combinations. However, since a single transition only changes the history for the current state $s$ and action $a$, the inference network will be affected only by the corresponding term in the full ELBO. Therefore, optimizing the entire ELBO at every step is equivalent to taking a gradient step with respect to the single ELBO in \ref{eqn:cmc_single_elbo} for the relevant $s$ and $a$. The derivation of the ELBO objective in \ref{eqn:cmc_single_elbo} and the full training algorithm for this model are included in Appendix \ref{appendix:cmc_deriv} in the supplementary material. 

\subsubsection{Action}
Our framework relies on uncertainty minimization for action selection. In this setting, our model selects actions that lead to the greatest reduction in its uncertainty about the inferred transition distributions $z_{s, a}$. This can be done by planning over long horizons \cite{active_inference_process_theory} or by using value iteration techniques \cite{little_sommer}. However, to keep our model simple and more computationally efficient, we employ a greedy approach with a simple heuristic that guides the model towards states with greater uncertainty. Despite this, our model is still able to achieve faster and more efficient exploration compared to random action selection. At each time step, the agent evaluates each action based on the following uncertainty reduction score
\begin{equation}\label{eqn:cmc_score}
    \text{Uncertainty Reduction}(a) = H\Big(q_\phi(z_{s, a}|\bm{h}_{s, a})\Big) - \mathbb{E}_{s'\sim z_{s, a}|\bm{h}_{s, a}}\left[H\Big(q_\phi(z_{s, a}|\bm{h}_{s, a}+s')\Big)\right]
\end{equation}
which is an adaptation of Equation \ref{eqn:uncertainty_red} to this specific application. Overall, Equation \ref{eqn:cmc_score} represents the expected reduction in uncertainty for action $a$ over a single step and a single transition distribution. To guide the agent towards future uncertain states, we add the following heuristic that describes the expected future uncertainty about the transition distributions in the next state
\begin{equation}\label{eqn:efu}
    \text{Expected Future Uncertainty} (a) = \mathbb{E}_{s'\sim z_{s, a}|\bm{h}_{s, a}}\left[\sum_{a'\in\mathcal{A}} H\Big(q_{\phi}(z_{s', a'}|\bm{h}(s', a'))\Big)\right]
\end{equation}
The final score maximized by the agent is the sum of the uncertainty reduction in \ref{eqn:cmc_score} and expected future uncertainty in \ref{eqn:efu}. 

\subsection{Active Vision Model}
\label{section:active_vision_model}
\subsubsection{Task}
\label{section:av_task}

\begin{figure}
\centering
    \begin{subfigure}[b]{0.35\textwidth}
        \includegraphics[width=\textwidth]{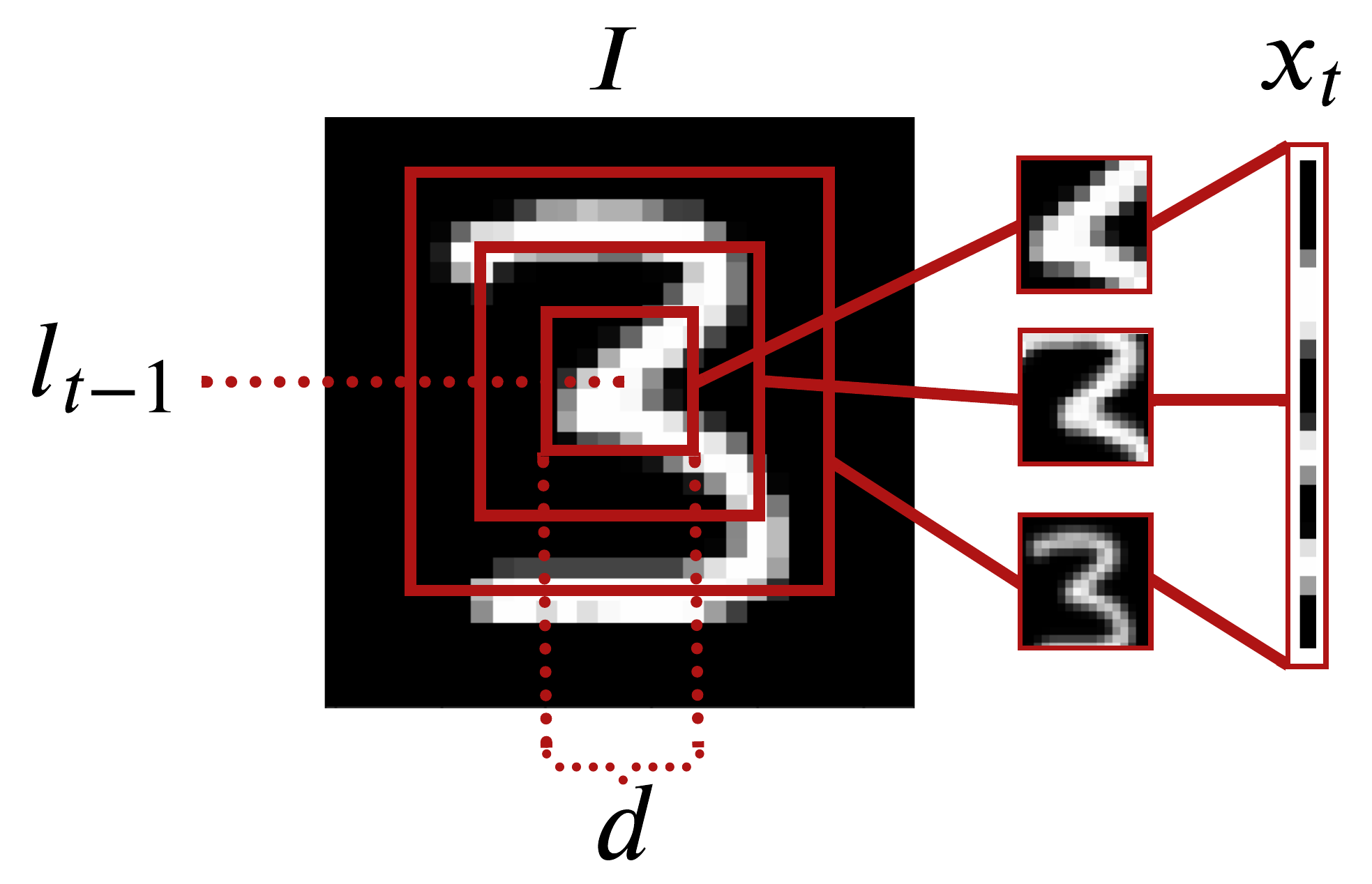}
        \caption{}
        \label{fig:foveation}
    \end{subfigure}
    \;\;\;
    \begin{subfigure}[b]{0.4\textwidth}
        \includegraphics[width=0.3\columnwidth]{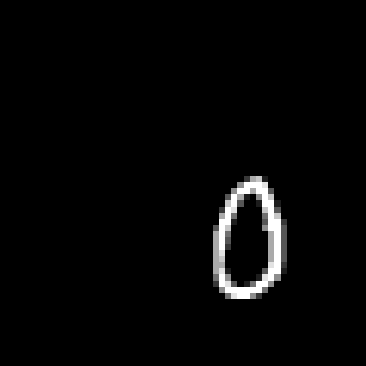}\;\includegraphics[width=0.3\columnwidth]{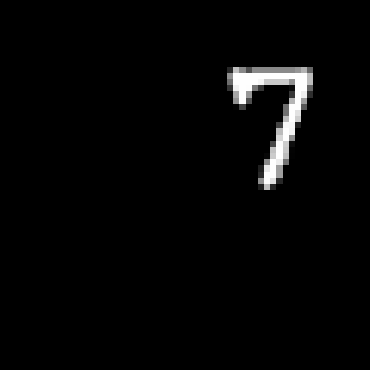}\;\includegraphics[width=0.3\columnwidth]{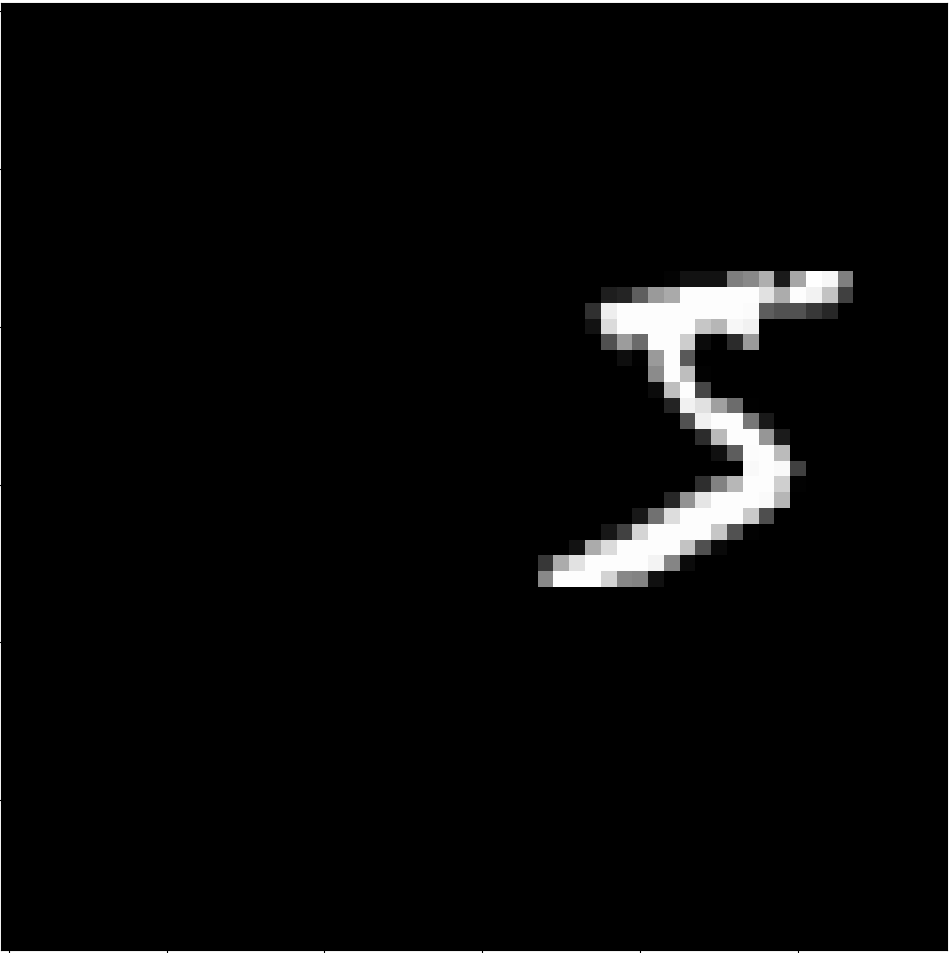}
        \vskip 0.37in
        \caption{}
        \label{fig:translated_samples}
    \end{subfigure}

    \caption{(a) Foveation setup for the bandlimited sensor in the active vision task. (b) Examples from the translated MNIST dataset used in our evaluations.}
    \label{fig:av_fov_samples}
\end{figure}

Next, we apply our model to the more complex task of active vision, where the state and action spaces are continuous and the state space is high-dimensional. In this task, the model explores a hidden image through a sequence of fixations. Each fixation yields a sample of the image at the fixation location. The size of this sample corresponds to the size of the model's fovea and is always less than the size of the input image. Furthermore, `foveated' samples can be extracted by the process illustrated in Figure \ref{fig:foveation}. Specifically, let $l_{t-1}$ denote the location of the fixation that generates the sample $x_t$ from the input image $I$. We use normalized coordinates so $l_{t-1} \in [-1, 1] \times [-1, 1]$, with $(-1, -1)$ corresponding to the top left corner of the image. Let $d$ denote both the height and width of the model's fovea and let $N_{fov}$ denote the number of foveation patches (the number of red squares in Fig. \ref{fig:foveation}). We first extract $N_{fov}$ patches of increasing size, all centered at $l_{t-1}$. We then downsample all patches so they all have the same size $d\times d$. The patches are then flattened and concatenated to generate $x_t$, which is the input to the model. Note that this is the same setup used in \cite{mnih_ram}.

Similarly to the CMC setting, our model is trained for active vision in the absence of an extrinsic training signal, such as classification loss. However, we show that the representations learned can yield high accuracy in a downstream classification task. It is important to note that when our model is trained to perform classification, the perception and action components are still not trained with the classification loss. Rather, the gradients from the classification loss are used to update only a separate feedforward decision network, which receives as input the internal representations of the perception model.

\subsubsection{Perception}
The active vision perception model is based on a simple hierarchical two-level generative model that reflects the structure of the task. This model is shown in Figure \ref{fig:av_model}. Let $I$ denote the image presented on a given trial. The higher level of the perception model encodes a single abstract representation $s$ which may reflect high-level properties of the class to which $I$ belongs. For example, $s$ can represent the number of lines and circles and the spatial correlations specific to drawing a certain digit. The lower level contains individual units whose activations are entirely driven by sensory input. From a biological perspective, these units correspond to neurons in the primary visual cortex whose receptive fields overlap with the spatial span of the fovea. At each time step $t$ and for a given sensory input $x_t$, we denote the activities of these neurons by $z_t$. From a generative perspective, $z_t$ can contain information about the lower level properties of image $I$ at a given location $l_{t-1}$. These properties, for example, can include stroke width, style, etc. 

Using the chain rule and the fact that the $\{x_{1:T}\}$ and $s$ are conditionally independent given $\{z_{1:T}\}$, the full variational posterior can then be factorized as a product of two variational posteriors,
\begin{equation}\label{eqn:av_var_posterior}
q(z_{1:T}, s|x_{1:T}, l_{0:T-1}) = q_1(z_{1:T}|x_{1:T}, l_{0:T-1}) \times q_2(s|z_{1:T}, l_{0:T-1})
\end{equation}

For notational simplicity, we will often omit the conditioning in the variational posteriors, e.g. use $q(z_{1:T}, s)$ to refer to $q(z_{1:T}, s|x_{1:T}, l_{0:T-1})$. The corresponding ELBO, derived in Appendix \ref{appendix:av_deriv}, is found to be \begin{equation}\label{eqn:av_elbo}
    \mathcal{L}_{ELBO} = \sum_{t=1}^{T} \mathbb{E}_q[\log{p(x_t|z_t)}] - \mathbb{E}_q\Big[\dfrac{p(s)}{q_2(s|z_{1:T}, l_{0:T-1})}\Big] - \sum_{t=1}^{T} \mathbb{E}_q\Big[\dfrac{p(z_t|s, l_{t-1})}{q_1(z_t|x_t, l_{t-1})}\Big]
\end{equation}
Note that the loss function does not depend on the entire image except at the locations sampled and viewed by the model. This maintains consistency with the natural setting, where the agent's perception does not encompass the entire image and so it cannot (as a whole) be used for training. Throughout our experiments, we assume the prior over $s$ to be a standard Gaussian. We also assume the likelihood distributions, $p(x_t|z_t)$ and $p(z_t|s, l_{t-1})$, as well as the variational posteriors, $q_1$ and $q_2$, to be Gaussian with means and variances parameterized by feed-forward neural networks. The full architecture of the model with these networks is shown in Figure \ref{fig:av_model} and is described below.

\paragraph{Network Architecture}
\label{subsec:net_arch}
The perception architecture consists of two variational autoencoders, one for each posterior in Eq.\ref{eqn:av_var_posterior}. The encoders and decoders for both VAEs are simple feedforward networks. The following is a description of each model component. 

\textbf{Lower-level VAE}: At each time step $t$, the model receives an observation $x_t$ which, together with the corresponding location $l_{t-1}$, is passed through an encoder network to infer the posterior over sensory representations $q_1(z_t)$. Let $f_{enc,\;\mu}^{(1)}(x_t, l_{t-1})$ and $f_{enc,\;\sigma}^{(1)}(x_t, l_{t-1})$ denote the outputs of the lower-level encoder network at time $t$. In our experiments, we assume $q_1(z_t)$ is an isotropic Gaussian $\mathcal{N}(z_t|\mu^{z}_{t}, \sigma^{z}_{t}I)$, where 
\begin{align}
    \mu^{z}_{t} =&\; f_{enc,\;\mu}^{(1)}(x_t, l_{t-1}) \\ 
    \sigma^{z}_{t} =&\; \exp\Big(\frac{1}{2}f_{enc,\;\sigma}^{(1)}(x_t, l_{t-1})\Big)
\end{align}

The lower-level VAE decoder network takes sensory representations $z_t$ and outputs the likelihood distribution $p(x_{t}|z_t)$. We assume this distribution is Gaussian $\mathcal{N}(x_t|\hat{x}_t, I)$, where $\hat{x}_t$ is the output of the decoder. 

\textbf{Higher-level VAE}: At the end of a fixation sequence, the higher-level encoder network receives the sum of past sensory representations and uses it to infer the posterior over abstract representations $q_2(s)$. Similar to $q_1(z_t)$, we assume $q_2(s)$ is an isotropic Gaussian $\mathcal{N}(s|\mu^{s}, \sigma^{s}I)$ parameterized by the output of the higher-level encoder $f_{enc}^{(2)}(h_T)$. The decoder network at this level receives an abstract representation $s$ and a query location $l_{t-1}$, and predicts a distribution over the corresponding lower-level representations $p(z_t|s, l_{t-1})$.

\paragraph{Generative Mechanism} We can generate new data from the model as follows. First, we sample an abstract representation $s$ from a standard Gaussian distribution. Then, we pick a query location $l'$ from the interval $[-1, 1] \times [-1, 1]$. Then, we pass $s$ and $l'$ through the higher-level decoder $f^{(2)}_{dec}$ which outputs a distribution $p(z')$. We sample $z'$ from this distribution and pass it through the lower-level decoder $f^{(1)}_{dec}$ which outputs an observation $x'$ that is the same size as the model's retina.

\subsubsection{Action}
Adapting the uncertainty reduction objective for action selection in active vision gives the following value function for a given action (or fixation location) $l_t$
\begin{equation}\label{action_value}
    V(l_t|x_{1:t}, l_{0:t-1}) := H(s|x_{1:t}, l_{0:t-1}) - \mathbb{E}_{p(x_{t+1}|s,l_{t})}\Big[H(s|x_{1:t+1}, l_{0:t})\Big],
\end{equation}

where $H(\cdot)$ denotes the Shannon entropy. Intuitively, $V(l_t)$ quantifies how much information the agent expects to gain as a result of observing the input image at location $l_t$. Therefore, a good estimate of information gain depends on how accurate the agent's generative model is. The objective in \ref{action_value} is intractable because it requires evaluating the posterior over $s$ for all possible observations $x_{t+1}$ in a continuous space. So, instead, we compute an approximation of it using a Monte Carlo (MC) sampling approach. First, we independently draw $K$ samples from the agent's likelihood model $p(x_{t+1}|s,l_t)$. The approximate expected entropy can then be computed as the average over the $K$ samples, which yields the approximate value
\begin{equation}\label{eqn:av_approx_value}
    \tilde{V}(l_t|x_{1:t}, l_{0:t-1}) \approx H(s|x_{1:t}, l_{0:t-1}) - \dfrac{1}{K} \sum_{k=1}^{K} H(s^{(k)}|x_{1:t}, x_{t+1}^{(k)}, l_{0:t}),
\end{equation}
where $x^{(k)}_{t+1}$ denotes the $k^{th}$ sample and $s^{(k)}$ denotes the corresponding updated posterior estimate. To make our model end-to-end differentiable, we use a neural network to select fixation locations that maximize the value in Equation \ref{eqn:av_approx_value}. This action network can be trained with gradient descent because all the terms in \ref{eqn:av_approx_value} are computed from the neural networks in the perception model, which receive fixation locations as part of their input. Therefore, it's possible to compute the gradients of \ref{eqn:av_approx_value} with respect to the output of the action network, which is  then trained to output actions with high informational value. For all experiments, the action network we use is a simple two-layer feedforward network which receives as input the current estimate of the state $s$ and outputs the mean of a Gaussian distribution over fixation locations. The standard deviation of this distribution is a fixed hyperparameter that we specify. The agent chooses a fixation location by sampling from the output distribution of the action network. Algorithm \ref{alg:bas_av} describes our differentiable approach for selecting continuous actions with uncertainty reduction in active vision tasks.

\begin{algorithm}
    \caption{Bayesian Action Selection in Active Vision}
    \label{alg:bas_av}
\begin{algorithmic}
   \STATE {\bfseries Input:} observations $x_{1:t}$, locations $l_{0:t-1}$, perception model $F$, action network $\psi_t$, number of MC samples $K$
   \STATE $q_t(s) = F.Encode(x_{1:t}, l_{0:t-1})$
   \STATE $l_t = \psi_t(\mathbb{E}[q_t(s)])$
   \STATE $p(x_{t+1}) = F.Decode(q_t(s), l_t)$
   \STATE Draw $K$ samples from $p(x_{t+1})$
   \FOR{$k=1$ {\bfseries to} $K$}
   \STATE $q_{t+1}^{(k)}(s) = F.Encode(x_{1:t}, x_{t+1}^{(k)}, l_{0:t-1})$
   \ENDFOR
   \STATE $\tilde{V}(l_t) = H(q_t(s)) - \dfrac{1}{K}\sum_{k=1}^{K} H\Big(q_{t+1}^{(k)}(s)\Big)$
   \STATE Update action network parameters using gradient descent on $\tilde{V}$: $\psi_{t+1} = \psi_{t} + \mu \nabla_{\psi_{t}}\Tilde{V}(l_t)$
   \STATE {\bfseries return:} selected action $l_t$ and updated action network $\psi_{t+1}$
\end{algorithmic}
\end{algorithm}

\section{Experimental Procedures}
\label{experiments}
\subsection{Controllable Markov chains}
We first test our model in the CMC setting and compare its performance to two baselines: random exploration and visitation-count-based Boltzmann exploration, where the probability of a transition is inversely proportional to its visitation count. More precisely, during Boltzmann exploration, actions are sampled from the distribution
\begin{equation}
    \pi(a_t) = \dfrac{\exp(-\frac{1}{\tau} \sum_{s'} \bm{h}_{s_t, a_t, s'})}{\sum_{a}\exp(-\frac{1}{\tau}\sum_{s'}\bm{h}_{s_t, a, s'})},
\end{equation}
where $\tau$ is a temperature parameter that is linearly annealed from 1.0 to 0.1 throughout the episode. For all agents, however, the perception model is used to learn the underlying distributions from observations. To quantify how well the model's learnt distributions approximate the true distributions in the environment, we use the measure of missing information ($I_{M}$),
\begin{equation}
    I_{M}(\mathcal{P}||\hat{\mathcal{P}}) = \sum_{s\in \mathcal{S}, a\in\mathcal{A}} D_{KL}\Big(p(:|s, a)||\hat{p}(:|s, a)\Big)
\end{equation}
where $p$ and $\hat{p}$ are the true and learnt distributions, respectively. Note that this is the same measure used in \cite{little_sommer}. To show that the perception model is indeed able to learn the true distributions in a CMC, we test it in the simple environment of Dense Worlds used in \cite{little_sommer}. In this environment, there are 10 states and 4 actions. For each state-action combination, a transition distribution is drawn from a Dirichlet distribution with the concentration parameter $\alpha = 1$. This environment tests only the perception model, since it is simple enough that random action selection can perform well if run for a sufficient number of steps. Then, to evaluate our action model, we test our agent versus a randomly-exploring agent in $6\times 6$ maze environments. The experiment settings as well as architecture and hyper-parameter specifications are included in Appendix \ref{appendix:cmc_hp}.

\subsection{Active vision}
\label{section:av_settings}

We tested our active vision model on multiple image datasets, including MNIST \citep{mnist_dataset}, fashion MNIST \citep{fmnist_dataset}, and grayscale CIFAR-10 \citep{cifar_dataset}. First, we test the model's ability to produce meaningful images by generating and combining small patches at different locations. The model's ability to do that reflects an implicit understanding of the spatial relationships between different locations on a given image. Second, despite the model being trained with unsupervised objectives, we tested its representations on a downstream image classification task, where only a separate decision network is trained with the supervised classification loss. For these experiments, we let the model observe a single patch of size $8 \times 8$ at each fixation location. We used a maximum of three active fixations. The dimensionalities of the latent variables $z_{t}$ and $s$ were 32 and 64, respectively.

We test the model's ability to build translation invariant representations using the translated MNIST dataset. This dataset consists of $60 \times 60$ images with a handwritten digit placed at a random location in the image. Examples are shown in Figure \ref{fig:translated_samples}. On these experiments, we let the model observe a foveated sample with three patches ($N_{fov} = 3$) all downsampled to size $12 \times 12$ pixels. We allowed the model a maximum of four active fixations for these experiments.

The full specification of hyperparameters and settings for the active vision experiments is included in Appendix \ref{appendix:av_hp}. For the image classification tasks, we evaluate four variants of our model to investigate the role of each component. Below is a description of each variant.

\textbf{BAS + Perception} This is our full proposed model. The fixation locations are selected using BAS and the internal representation $s$ of the perception model is used as the input to the decision network. 

\textbf{Random + Perception} Here, fixation locations are selected randomly. The input to the decision network is still the abstract state $s$ inferred by the perception model.

\textbf{BAS + RNN} Here, the fixation locations are selected with BAS. However, instead of using the internal representations $s$ as input to the decision network, we use a separate RNN that integrates the collected observations $x_{1:T}$ and passes its hidden state vector to the decision network. To ensure a fair comparison between this variant and the perception-based variants, the size of this separate RNN is always the same as the dimensionality of the abstract latent variable $s$. Note that, in this case, the perception model is still involved in the action selection process since BAS determines action scores based on the entropy of the latent state distribution. 

\textbf{Random + RNN} This is the same as BAS + RNN but we replace BAS with a random selection of fixation locations.

\section{Results}
\label{results}
\subsection{Efficient exploration in CMCs}
To show that our perception model is able to learn the underlying transition distributions of CMCs, we run it for 2000 time steps in the Dense Worlds environment. We then compute the differences between the distributions learned by the agent and the true environment distributions. To see the effect of our exploration strategy, the same analysis is done for a randomly-exploring agent that uses the same perception model. Figure \ref{fig:prob_diffs} in the Appendix shows that our model is indeed able to learn good estimates of the environment's transition distributions. These estimates are better when data are collected using BAS, which shows that our agent collects data that are particularly beneficial for the learning progress of the perception model. 

\begin{figure}
    \centering
    \begin{subfigure}[c]{0.59\textwidth}
        \includegraphics[width=\textwidth]{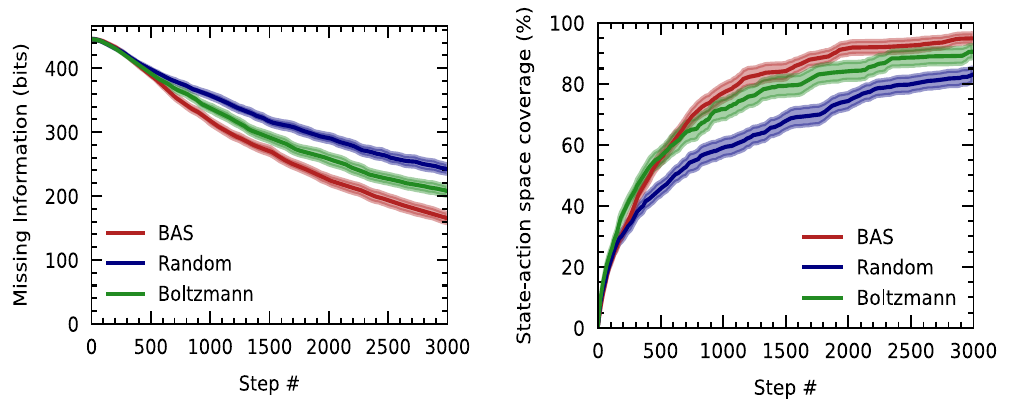}
        \caption{}
        \label{fig:maze_results_6x6}
    \end{subfigure}
    \centering
    \;
    \begin{subfigure}[c]{0.39\textwidth}
        \includegraphics[width=\textwidth]{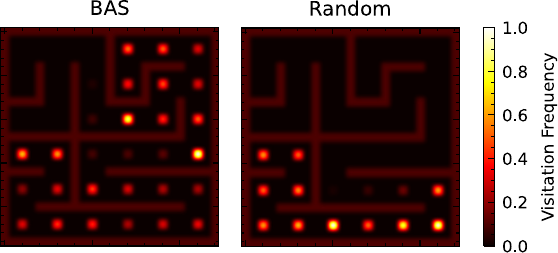}
        \begin{minipage}{20cm}
            \vfill
        \end{minipage}
        \caption{}
        \label{fig:maze_hms}
    \end{subfigure}
    \caption{Results of our active exploration model in the maze environment. (a) Missing information and percent state-action space coverage for a 6x6 maze. (b) example visitation frequency maps for a $6 \times 6$ maze explored by BAS (our model) versus a random exploration strategy. Both agents were allowed to run in the environment for a 1000 time steps. Visitation frequencies are normalized by the maximum visitation frequency in each case.}
    \label{fig:maze_results}
\end{figure}

Next, we test our model's ability to explore more complicated maze environments. We test three agents all using our perception model but employing different exploration strategies: BAS, random action selection, and Boltzmann exploration. The test environment was a $6\times 6$ maze and the agents were allowed to navigate for 3000 time steps. We also performed the same tests in larger mazes and obtained similar. Figure \ref{fig:maze_results_6x6} shows that exploration using BAS results in significantly faster reduction in missing information compared to the other two baselines, despite BAS and Boltzmann achieving similar state-action space coverage performance. Figure \ref{fig:maze_results_6x6} also shows that, while the BAS explorer covers the state-action space more quickly than the Boltzmann explorer, it does so more efficiently than the random explorer. To see this more clearly, we visualize the visitation frequencies of the BAS and random explorers as heat maps that show how much time an agent spends in each state (Figures \ref{fig:maze_hms} and \ref{fig:more_maze_heatmaps_6x6}). 

\subsection{Probing the generative model of active vision}
Predictive coding posits that the brain learns and maintains a generative model of the world that allows it to make good predictions about the environment. Additionally, since we assume that action selection relies on this model, the optimality of our actions depends directly on the quality of our generative models. In this section, we investigate how good our trained model is at generating new patches of images and inferring the underlying states from sequences of random fixations. 

\begin{figure}
    \;\;\;\;\;
    \centering
    \begin{subfigure}[c]{0.5\textwidth}
    \centerline{\includegraphics[width=0.3\textwidth]{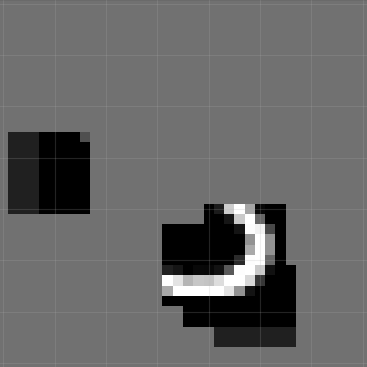}\;\;\;\includegraphics[width=0.3\textwidth]{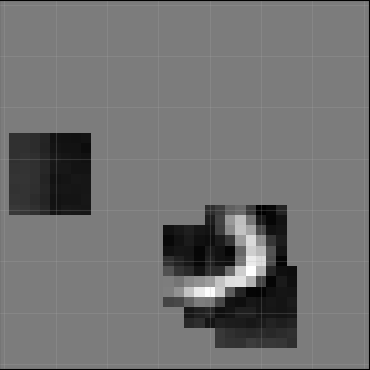}\;\;\;\includegraphics[width=0.3\textwidth]{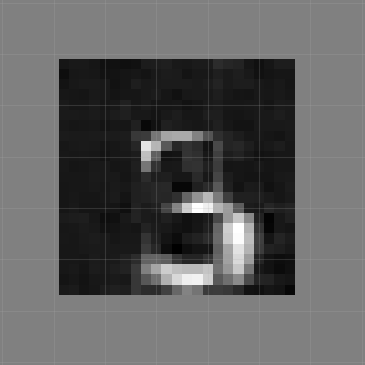}}
    %\centerline{}
    \centerline{\includegraphics[width=0.3\textwidth]{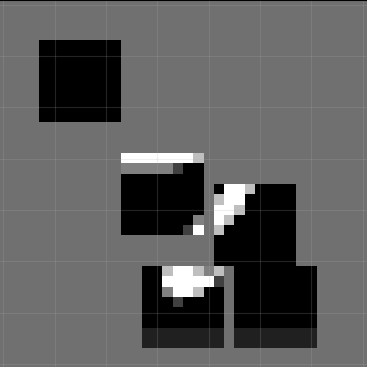}\;\;\;\includegraphics[width=0.3\textwidth]{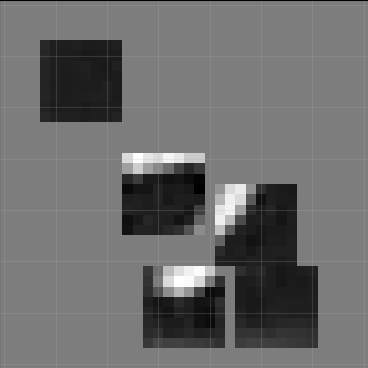}\;\;\;\includegraphics[width=0.3\textwidth]{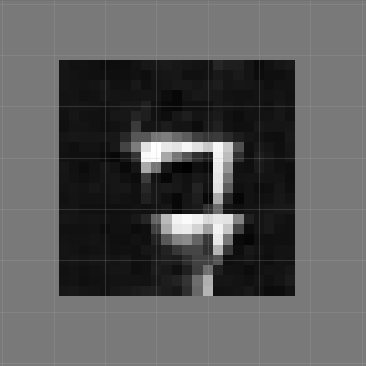}}
    \centerline{\includegraphics[width=0.3\textwidth]{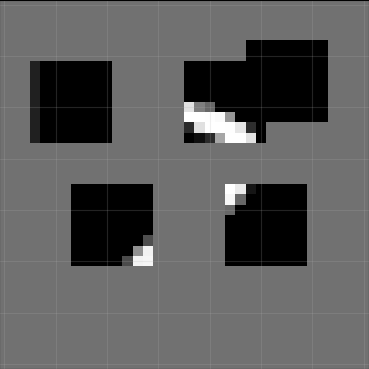}\;\;\;\includegraphics[width=0.3\textwidth]{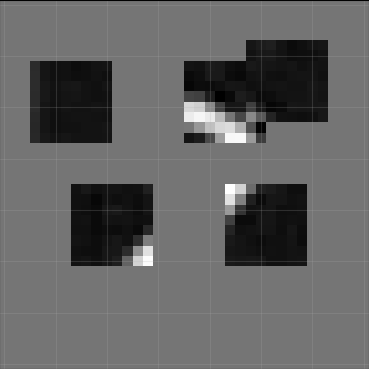}\;\;\;\includegraphics[width=0.3\textwidth]{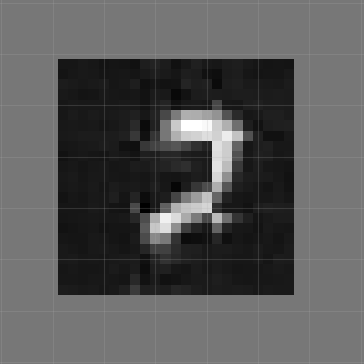}}
    \caption{}
    \label{fig:gen_tests}
    \end{subfigure}
    \hfill
    \begin{subfigure}[c]{0.45\textwidth}
    \vskip -0.02in
    \centerline{\includegraphics[width=0.33\textwidth]{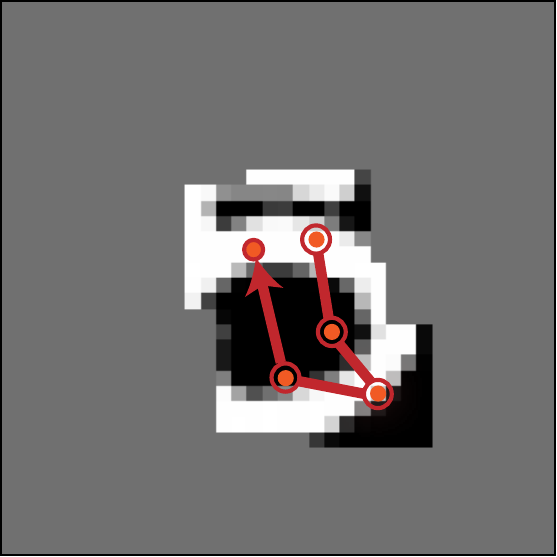}\;\;\;\includegraphics[width=0.33\textwidth]{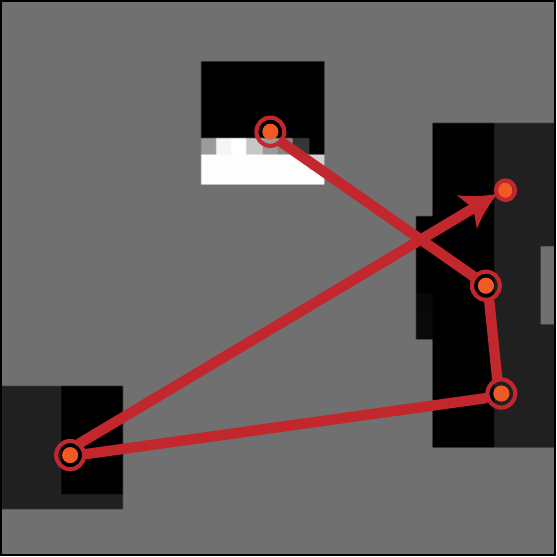}}
    \centerline{\includegraphics[width=0.33\textwidth]{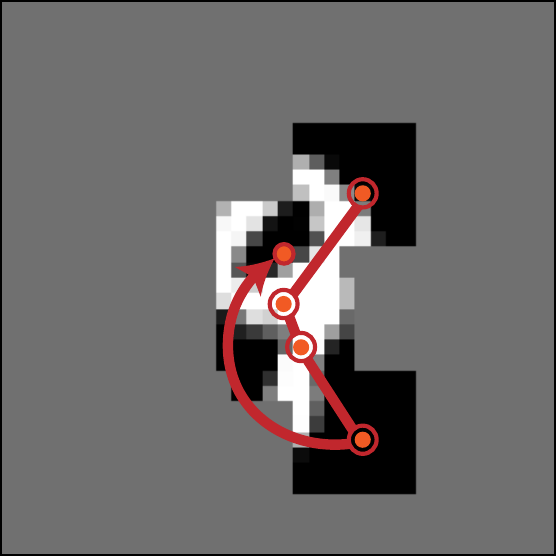}\;\;\;\includegraphics[width=0.33\textwidth]{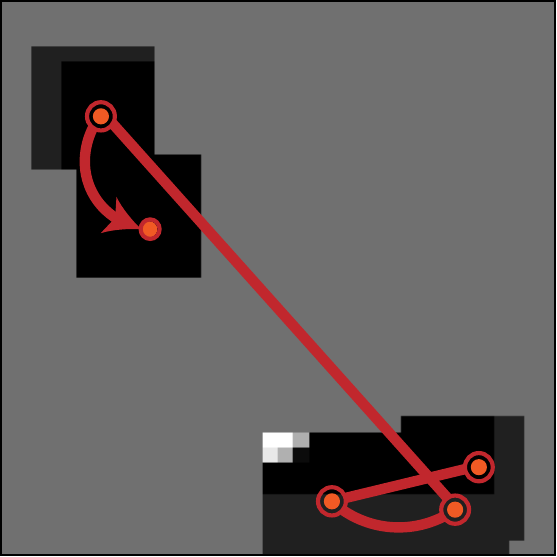}}
    \centerline{\includegraphics[width=0.33\textwidth]{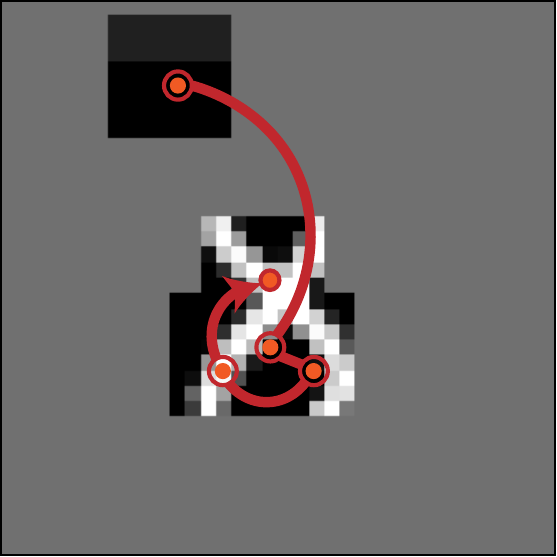}\;\;\;\includegraphics[width=0.33\textwidth]{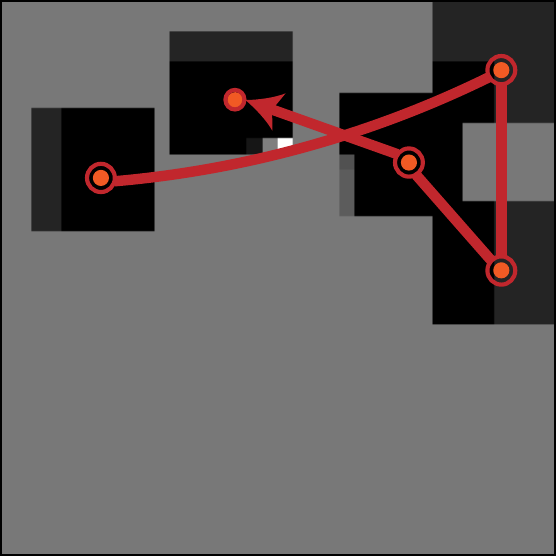}}
    \caption{}
    \label{fig:fix_seqs}
    \end{subfigure}
    \caption{Demonstrating the generative ability of the perception model and its influence on action selection. (a) Original patches of input images (left) and their reconstructions (middle). After the model infers an abstract representation, it is able to generate an imagined digit at the unobserved locations (right). (b) Fixation sequences generated using BAS (left column) and random strategies (right column).}
    \label{fig:probing_gen_model}
\end{figure}

Figure \ref{fig:gen_tests} shows examples of trials in which a random sequence of patches is given to the perception model. At the end of the sequence, the model infers the abstract state $s$ that might underlie the given observations. From its estimate of $s$, it computes reconstructions of each observed patch. Additionally, we can generate unobserved patches from this inferred estimate by querying the decoder networks at different locations in space. As seen in Figure \ref{fig:gen_tests}, when the generated patches at the nine central locations are put together, we get a meaningful image that corresponds to what the model \textit{imagines} the underlying digit is. This is interesting since the entire image at once is never observed by the model, nor is it used in the training losses. A similar effect is observed when testing the model on CIFAR-10 images (Figure \ref{fig:cifar_samples}), although the generated images tend to capture global statistics rather than local details due to the simplicity of our architecture, which utilizes only feedforward networks. These results demonstrate that the model successfully learns the spatial relationships between patches corresponding to individual image categories in a completely unsupervised manner, which explains its superior performance during classification later on. 

\subsection{Interaction between the perception and action models}
One of the benefits of our framework is its modularity (perception and action components) which allows us to look at how each component affects the other.
First, we look at how the representations learned by the perception model affect what actions are selected. In the centered MNIST dataset, the most informative location about the category of the image is the center. This is reflected in the representations of the perception model, which is able to produce meaningful digits by generating and combining individual patches. Therefore, a strategy that minimizes uncertainty would ideally choose to fixate at the center most of the time. Figure \ref{fig:fix_seqs} shows that this is exactly the case. When we compare fixation sequences selected by BAS versus a random strategy, the BAS strategy almost always chooses the center as its second fixation location after the initial random fixation. This shows that the statistical regularities in the environment are reflected in the behavior of the action model. 

Second, we study how selecting actions with BAS affects the latent representations developed through perception. We trained two perception models on data collected with random fixations and data collected with BAS. Then, we presented both models with data collected using BAS and examined the resulting representations in their latent spaces. We use Principal Component Analysis (PCA) and t-SNE \citep{tsne} to visualize these representations in 3D space, as shown in Figures \ref{fig:perception_pca} and \ref{fig:perception_tsne} in the appendix. We also quantify the quality of the representations with respect to image categories by estimating the mutual information \cite{mi_est1, mi_est2} between the three PC features and image labels, as shown in Figure \ref{fig:mutual_info}. 

%Results show the BAS-trained model is able to learn much better representations that are well-clustered in the latent space, as opposed to the model trained with random fixations. 

\subsection{Performance on downstream image classification}
\begin{figure}
    \centering
    \begin{subfigure}[c]{0.35\textwidth}
        \includegraphics[width=\textwidth]{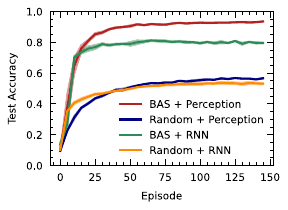}
        \caption{}
        \label{fig:mnist_perf}
    \end{subfigure}
    \;
    \begin{subfigure}[c]{0.35\textwidth}
        \includegraphics[width=\textwidth]{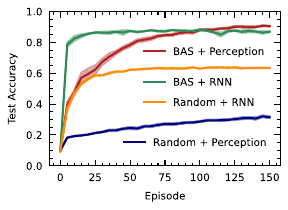}
        \caption{}
        \label{fig:trans_mnist_perf}
    \end{subfigure}
    \;
    \begin{subfigure}[c]{0.26\textwidth}
    \vskip -0.05in
    \includegraphics[width=\textwidth]{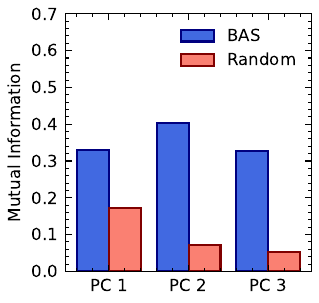}
    \vskip 0.1in
    \caption{}
    \label{fig:mutual_info}    
\end{subfigure}
    \caption{Performance on (a) the centered MNIST dataset (N = 5 random seeds) and (b) the translated MNIST dataset (N = 5 random seeds). Error bars indicate SEM. (c) mutual information between the latent representations' PC projections and class labels.}
    \label{fig:classification_perf}
\end{figure}

We demonstrate our model's ability to perform image recognition using the unsupervised representations it learns through free visual exploration. Figure \ref{fig:mnist_perf} shows performance on the centered MNIST dataset. Performance on fashion MNIST is reported in Figure \ref{fig:fmnist_results} of the appendix. In general, our BAS strategy yields better performance than a random action selection strategy. Furthermore, when the internal states of the perception model are used as input to the decision network, the classification accuracy is higher compared to using a separate RNN that integrates previous observations, indicating that the learned representations are more informative about the data. 

In the more difficult task of classifying translated digits, the performance generally gets worse. However, we can still see that our BAS strategy outperforms a random exploration strategy. The representations of the perception model, however, do not seem to offer more benefit than a regular RNN. One reason for this might be the absence of a statistical regularity in the locations of digits. Therefore, encoding abstract representations in an individual state $s$ may not be sufficient since other hidden states, such as digit location, affect the generative process. Nevertheless, the model is able to learn informative representations as evidenced by the effectiveness of BAS in selecting fixation locations. Note that, in the case of translated MNIST, using the foveation method described in Section \ref{section:av_task} results in a larger area being observed at each location, albeit with lower resolution towards the periphery. Therefore, it is possible for the model to accumulate observations of all parts of the digit during the fixation process. This may explain why the performance of RNN-based methods improves on translated compared to centered MNIST.

\subsection{Faster training and generalization with Bayesian Action Selection}
\label{sec:data_eff}

An important feature of our approach is that it can be trained in a completely unsupervised manner to explore visual scenes and build generative representations of them. We asked if this feature can help improve the computational efficiency and training speed of a separate downstream classifier whose parameters are trained with the supervised classification loss. 

\begin{figure}
    \centering
    \begin{subfigure}[c]{0.4\textwidth}
    \includegraphics[width=\textwidth]{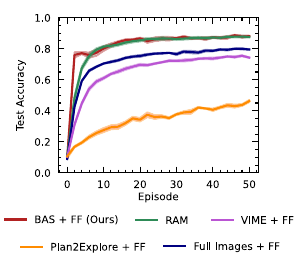}
        \caption{}
        \label{fig:training_speed}
    \end{subfigure}
    \;\;\;\;\;\;\;\;\;\;
    \begin{subfigure}[c]{0.37\textwidth}
        \vskip 0.12in
        \includegraphics[width=\textwidth]{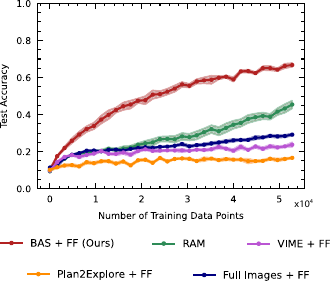}
        \vskip 0.12in
        \caption{}
        \label{fig:data_eff}
    \end{subfigure}
    \caption{Learning speed and data efficiency evaluations. (a) Comparison of the three cases described in the text in terms of their speed of training on the classification task with translated MNIST. Shaded error bars represent the SEM (n = 5 random seeds). (b) Data efficiency and generalization. We compare the same three methods in (a) in terms of their test performance during the first episode of supervised training, when the classification networks see the data for the first time. Each point on the plot represents the test score after observing only $x$ training data points for the first time. Shaded error bars represent the SEM (n = 5 random seeds).}
    \label{fig:efficiency_results}
\end{figure}

To test this, we look at the learning speed of a downstream classifier trained with full images (Full Images + FF) versus one trained on concatenated patches collected using BAS (BAS + FF). We also included three popular baselines from the RL literature in this evaluation: the Recurrent Attention Model (RAM) \citep{mnih_ram}, VIME \citep{vime}, and Plan2Explore \citep{p2e}. RAM is a popular method from the machine learning literature that is known to achieve high performance on the same task we consider here, while VIME and Plan2Explore are popular intrinsically-motivated exploration methods in RL that use information-theoretic measures similar to ours. We conducted these tests on the translated MNIST data because it is more complex and has a higher number of dimensions. In all test cases, the decision network used for classification had two hidden layers with a consistent number of units across all conditions. However, since our BAS strategy selects a few locations to observe on the full image, the total number of parameters trained with the supervised loss was approximately 50\% less for the model trained with BAS-collected data than for the model trained with full images. Figure \ref{fig:training_speed} shows that, in addition to having lower parameter complexity, our method learns much faster than all other baselines and achieves higher asymptotic performance than VIME, Plan2Explore, and Full Images + FF.

We also considered a comparison of the same methods described above in terms of their data efficiency. Specifically, we ask the following question: during the first supervised training episode, how many training examples does a model need to observe to reach a given performance on the test set? This question addresses issues of few-shot learning and fast generalization. Since our BAS strategy utilizes the perception model's abstract representations of the task, we hypothesized that it would lead to a higher test performance with fewer training examples. This is exactly what we find through our analysis, as shown in Figure \ref{fig:data_eff}. From these results, we see that our model is able to learn significantly faster from fewer training examples, highlighting the generalizability and effectiveness of the model's abstract representations in guiding action selection. A description of the experiments and hyperparameters used in these analyses is included in Appendix \ref{appendix:av_hp}.

\section{Discussion}
\label{discussion}
We developed a biologically inspired model of active sensing by combining two theories from neuroscience: predictive coding for perception and uncertainty minimization for action. Although these two theories have been utilized previously, our model incorporates them in a unique, scalable, and end-to-end framework, enabling flexible intrinsically driven exploration for embodied AI. Furthermore, the proposed model provides an approximate method for learning policies that optimize information gain in a differentiable manner, utilizing a deep generative model. We test this model in two sensorimotor tasks that integrate perception and action: 1) learning transition dynamics through pure exploration in discrete environments, and 2) learning unsupervised representations in active vision.

An important aspect of our approach is its generality. Specifically, it can be applied to any perception-action setting while only requiring the specification of a generative model that relates internal representations to sensory observations. By parameterizing this generative model with neural networks, we are able to compute uncertainty with respect to perceptual states in a differentiable manner, allowing the action and perception models to interact in an end-to-end fashion. We have shown how to instantiate this framework in the discrete setting of CMCs as well as the continuous setting of active vision. We note that the perception model in each case is a VAE, despite the different nature of each task and thus the different generative model. In both cases, the ELBO objective is used to learn a probabilistic relationship between latent states and observed variables. However, the action selection model is different due to the discrete nature of the CMC problem versus its continuous counterpart in the active vision task. In CMCs, actions are evaluated directly, while in active vision, a neural network is trained to output continuous actions. Nonetheless, action selection in each case still aims to maximize uncertainty reduction, complying with our general active exploration framework.

Our approach emphasizes the relationship between perception, action, and learning during exploratory behavior. The importance of this relationship shows up in the CMC and active vision settings. Specifically, we see that, in both cases, the quality of the representations learned by the perception model directly depends on the exploration strategy used to collect information from the environment. For example, in CMCs, we see that our BAS exploration strategy leads to faster and more efficient learning of the underlying transition distributions compared to Boltzmann and random exploration (Figure \ref{fig:maze_results_6x6}). This is despite BAS and Boltzmann performing somewhat similarly in terms of state-action space coverage, which shows that the uncertainty reduction objective leads to collecting observations that are particularly useful for the perception model to learn accurate representations of its environment. In active vision, we see a similar effect; when the perception model is trained on data collected with BAS (as opposed to random) exploration, the representations are well clustered in the latent space and contain more information about image categories (Figures \ref{fig:perception_pca} and \ref{fig:mutual_info}). These perception-action dependencies are captured so efficiently in our approach as a result of using a generative model with respect to which uncertainty can be measured. In contrast, RL exploration methods (e.g. \cite{vime, max, stadie2015incentivizing, pathak2017curiosity}) either develop approximate methods for information-based action selection or learn only the dynamics of a given MDP instance solely to aid in exploration. As a result, these methods are mainly concerned with efficient policies for action (exploration) without much emphasis on perception, i.e., learning useful representations of the environment \emph{through} exploration.  

% limitations and future directions 
In discrete-action settings such as CMCs, a potential limitation of our approach is that action scoring, which relies on enumeration, may not scale well to larger environments with bigger state-action spaces. Although directly evaluating actions is more accurate, we can still select actions with sufficient accuracy while improving scalability by using a neural network trained to minimize uncertainty, similar to the active vision model. Additionally, one potential concern (especially for the active vision setting) relates to how well this approach can be applied to more complex real-world datasets. In this work, we opted to keep our models as simple as possible (using only feedforward networks) to illustrate the advantages of our approach. It is, however, possible to implement this framework with more complex architectures, utilizing more advanced types of networks such as Convolutional Neural Networks, which may allow for applications in more complicated real-world problems. Such applications will be interesting to investigate as a future extension of this work.

\section{Code Availability}
\label{code}
The code and data to reproduce the results in this article are available \href{https://github.com/AbdoSharaf98/active-sensing-paper.git}{here}.

\section{Acknowledgement}
\label{acknowledgement}
This work was supported by a Sloan Research Fellowship in Neuroscience to H.C. and an NSF award \#2223811 to N.I..
We would also like to thank Rajesh Rao for helpful suggestions and comments on our manuscript.

\section{Author Contributions}
Conceptualization, A.S., N.I., and H.C.; Methodology, A.S., N.I., and H.C.; Software, A.S.; Formal Analysis, A.S.; Investigation, A.S., Resources; N.I. and H.C.; Writing - Original Draft, A.S.; Writing - Review \& Editing, A.S., N.I., and H.C.; Visualization, A.S.; Supervision, N.I. and H.C.; Project Administration, N.I. and H.C.; Funding Acquisition, N.I. and H.C.

\section{Declaration of Interests}
The authors declare no competing interests.

%%%%%%%%%%%%%%%%%%%%%%%%%%%%%%%%%%%%%%%%%%%%%%%%%%%%%%%%%%%%%%%%%%%%%%%%%%%%%%%%%%%%%%%%%%%%%%%%%%%%%%%%%%%%%%%%%%%%%%%%
\newpage
\appendix
\renewcommand{\thesection}{\Alph{section}}
\newcommand{\theHalgorithm}{S\arabic{algorithm}}
\renewcommand{\thefigure}{S\arabic{figure}}

\setcounter{figure}{0}

\section{Relationship between Predictive Coding and Variational Autoencoders}
\label{appendix:pc_vae}
\begin{wrapfigure}{R}{0.4\textwidth}
\vskip 0.1in
\begin{center}
\centerline{\includegraphics[width=0.2\columnwidth]{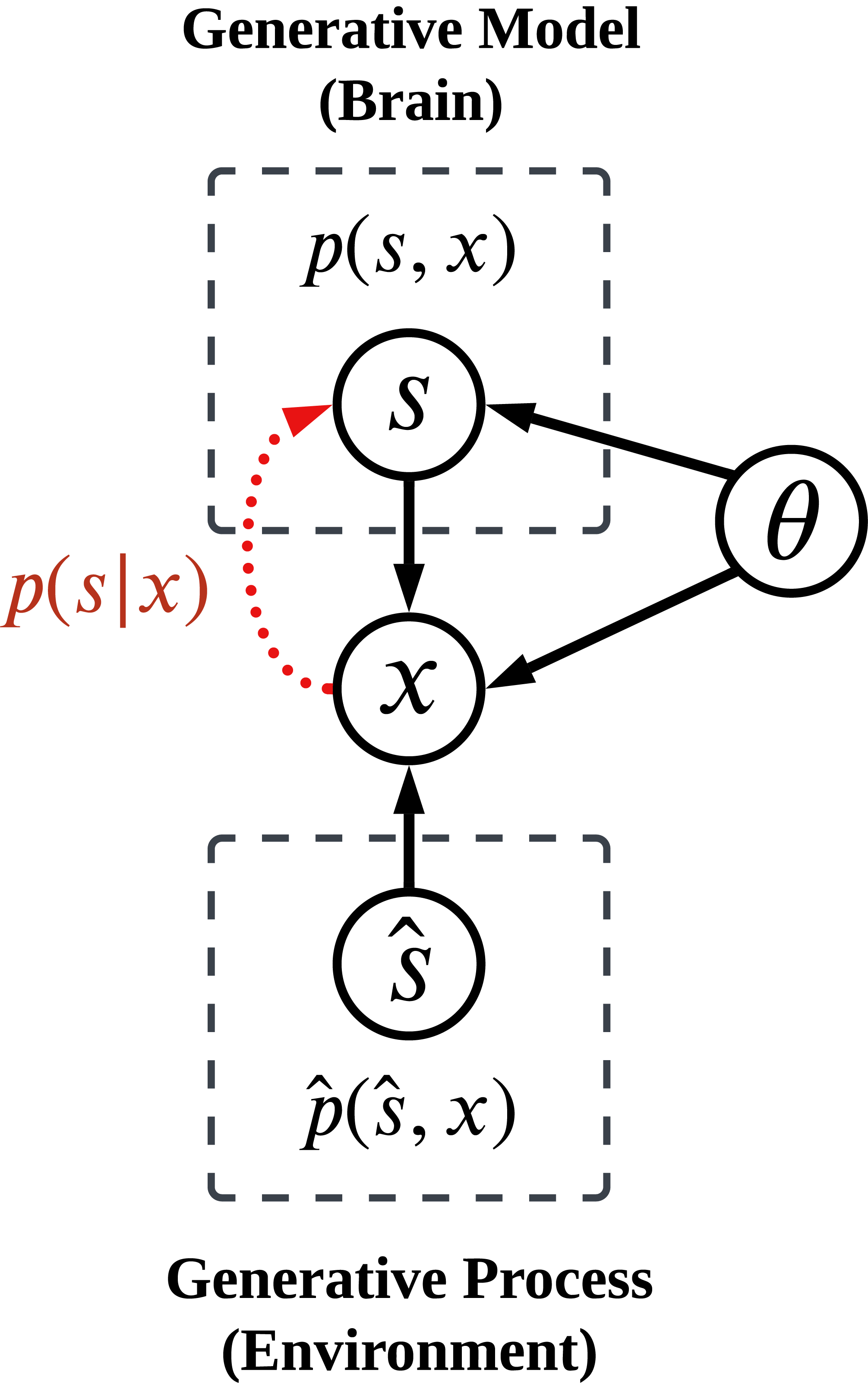}}
\caption{Simple two-layer hierarchical generative model, parameterized by $\theta$, which approximates the true distribution of the generative process giving rise to observations $x$. The predictive coding framework postulates that neural activities encode hidden state estimates, e.g. $s$. Thus, perception corresponds to inverting the generative model and inferring those neural activities (red arrow).}
\label{fig:pred_coding_gen_model}
\end{center}
\vskip -0.8in
\end{wrapfigure}

According to the predictive coding framework, the brain maintains a generative model of the world which approximates a mapping between observed sensory input and hidden states of the environment. This is illustrated in Figure \ref{fig:pred_coding_gen_model}. Perception, therefore, corresponds to inverting this model to infer hidden states, while learning corresponds to updating the parameters of this model based on prediction errors. Here, we outline the relationship between hierarchical predictive coding as presented in \cite{rao_ballard} and the framework of Variational Auto-encoders in machine learning [\cite{kingma_vae}]. A similar outline of this relationship is given in \cite{pred_theories} and \cite{marino_pc_vae} (with more details on the connections between theory and biology). 

To simplify the discussion, we assume the generative model consists of two hierarchical layers (an input layer and a sensory layer) as shown in Figure \ref{fig:pred_coding_gen_model}. In reality, the sensory areas in the brain contain many more hierarchical levels, and this discussion can be easily extended to multi-layer hierarchical generative models. 

The goal of inference is to find the best estimate $s^\star$ of the true hidden state $\hat{s}$ given an observation $x$. In the predictive coding framework, this is done by maximizing the posterior distribution $p(s|x)$, which is known as maximum a posteriori (MAP) estimation. Equivalently, we can maximize $\log{p(s|x)}$ since the log is a monotonic function in $p$. This problem can be formulated as follows
\begin{align}
    s^\star =&\; \arg\max_{s} \log{p(s|x)} \\
    =&\; \arg\max_{s} \log{\dfrac{p(x|s)p(s)}{p(x)}}\\
    =&\; \arg\max_{s} [\log{p(x|s)} + \log{p(s)}]\label{eqn:map_eq}
\end{align}
To perform this optimization, we adopt some assumptions about the likelihood distribution $p(x|s)$ and the prior on the hidden state $p(s)$. In their original implementation, Rao and Ballard assume the following parameterizations
\begin{align}
    p(x|s) =&\; \mathcal{N}(f(\bm{W}s); \sigma_x^2 \bm{I})\label{eqn:likelihood_eqn}\\
    p(s) =&\; \mathcal{N}(\mu_s; \sigma_s^2\bm{I})\label{eqn:prior_eqn}
\end{align}
where $\bm{W}$ is a weight matrix, $f(\cdot)$ is a non-linearity, and $\mathcal{N}(\mu, \sigma\bm{I})$ is an isotropic gassuian with mean $\mu$ and covariance $\sigma \bm{I}$. Without loss of generality, we can simplify this further by assuming the prior $p(s)$ is a standard gaussian, i.e. $\mu_s = \bm{0}$ and $\sigma_s^2 = 1$. Substituting this into equation \ref{eqn:map_eq}, we get 
\begin{align}
    s^\star =&\; \arg\max_s\; \log{\mathcal{N}(f(\bm{W}s); \sigma_x^2 \bm{I})} + \log{\mathcal{N}(0; \bm{I})}\\
    =&\; \arg\min_s\; \dfrac{1}{\sigma_x^2} \|x - f(\bm{W}s)\|_2^2 + \|s\|_2^2\label{eqn:pred_coding_obj}
\end{align}
Equation \ref{eqn:pred_coding_obj} is the predictive coding objective for the simple generative model in Figure \ref{fig:pred_coding_gen_model}. The first term in the objective is a reconstruction loss (or prediction error) and the second term is a regularization term which ensures that the inferred state $s^\star$  is consistent with the prior over $s$. To learn the parameters $\bm{W}$, the same objective is minimized with respect to $\bm{W}$ while fixing the inferred state $s$. 

The predictive coding formulation described above attempts to find a point estimate $s^\star$ which maximizes $p(s|x)$. An alternative is to find the full posterior distribution
\begin{equation}
    p(s|x) = \dfrac{p(x|s) p(s)}{p(x)} = \dfrac{p(x|s)p(s)}{\int_s p(x, s) ds}
\end{equation}
This is intractable to do exactly since it requires evaluating an integral over the continuous space of hidden states. Variational inference \cite{jordan_variational_methods} allows us to approximate the posterior $p(s|x)$ with some \textit{variational} posterior $q(s)$. Given a family of distributions $\mathcal{Q}$ defined over the space of hidden states $s$, we aim to find the distribution $q(s) \in \mathcal{Q}$ which minimizes the objective 
\begin{align}
    D_{KL}\Big(q(s)||p(s|x)\Big) =&\; \mathbb{E}_{s\sim q(s)}\Big[ \log{\dfrac{q(s)}{p(s|x)}}\Big]\\
    =&\; \mathbb{E}_{s\sim q(s)}\Big[ \log{\dfrac{q(s)p(x)}{p(x, s)}}\Big]\\
    =&\; \log{p(x)} - \mathbb{E}_{s\sim q(s)}\Big[ \log{\dfrac{p(x, s)}{q(s)}}\Big] 
\end{align}
The quantity $\mathbb{E}_{s\sim q(s)}\Big[ \log{\dfrac{p(x, s)}{q(s)}}\Big]$ is known as the evidence lower bound (ELBO) because, due to the non-negativity of the KL divergence, it forms a lower bound on the log evidence 
\begin{align}
    \log{p(x)} \geq \mathbb{E}_{s\sim q(s)}\Big[ \log{\dfrac{p(x, s)}{q(s)}}\Big] = \mathbb{E}_{s\sim q(s)}[\log{p(x|s)}] - D_{KL}\Big(q(s)||p(s)\Big)
\end{align}
Therefore, minimizing the KL term $D_{KL}\Big(q(s)||p(s|x)\Big)$ amounts to maximizing the ELBO since $\log{p(x)}$ does not depend on either $s$ or the parameters of the model. When we factor in the assumption in Equation \ref{eqn:likelihood_eqn}, the ELBO reduces to the objective 
\begin{align}
    -\mathbb{E}_{s\sim q(s)} \Big[\|x - f(\bm{W}s)\|_2^2\Big] - D_{KL}\Big(q(s)||p(s)\Big)\label{eqn:general_elbo}
\end{align}

To see the correspondence between the ELBO in Equation \ref{eqn:general_elbo} and the predictive coding objective in Equation \ref{eqn:pred_coding_obj}, note that the first term in Equation \ref{eqn:general_elbo} leads to the minimization of the reconstruction error, while the second term constrains the deviation of the posterior $q(s)$ from the prior $p(s)$. Finally, We can train neural networks to optimize the ELBO in equation \ref{eqn:general_elbo} using the framework of Variational Autoencoders \cite{kingma_vae}. 

\newpage
\section{Training of active exploration model in controllable Markov chains}
\label{appendix:cmc_deriv}

Algorithm \ref{alg:bas_cmc} describes the Bayesian Action Selection (BAS) algorithm in controllable Markov chains (CMCs) with discrete action and state spaces. The training algorithm for our full active sensor model (including perception) in CMCs is outlined in Algorithm \ref{alg:active_sensing_cmc}. After the model is trained for $T$ time steps in the environment, the learned transition distribution $\hat{p}(:| s, a)$ for a given state $s$ and action $a$ is found as the mean of the Dirichlet posterior output by the perception network at the end of training. That is,
\begin{align}
    q(z_{s, a}) =&\; \text{Dir}\Big( \phi_{T}(s, a, \bm{h}_{s, a})\Big)\\
    \hat{p}(:|s, a) =&\; \mathbb{E}_{q}[z_{s, a}] 
\end{align}

The ELBO objective in Equation 5 of the main text can be derived as follows.
\begin{align}
    \log{p(\mathcal{H})} =&\; \log{\int_{z}q(z|\mathcal{H})\dfrac{p(z, \mathcal{H})}{q(z|\mathcal{H}})\;dz}\\
    =&\; \log{\mathbb{E}_{q}\left[\dfrac{p(z, \mathcal{H})}{q(z|\mathcal{H})}\right]}\\
    \geq&\; \mathbb{E}_q\left[ \log{\dfrac{p(\mathcal{H}|z)p(z)}{q(z|\mathcal{H})}}\right]\label{eqn:jensen's_cmc}\\
    =&\; \mathbb{E}_q\left[ \log{p(\mathcal{H}|z)}\right] - D_{KL}\left(q(z|\mathcal{H})||p(z)\right) = \mathcal{L}_{ELBO}\label{eqn:elbo_cmc}
\end{align}
where Equation \ref{eqn:jensen's_cmc} follows from Jensen's inequality. In practice, we can multiply the KL term in \ref{eqn:elbo_cmc} by a scalar $\beta$ to control the balance between the regularity of the learned distributions and how well they explain past experiences \cite{beta_vaes}. 
\vfill
\begin{algorithm}[h]
   \caption{Bayesian Action Selection in Controllable Markov Chains}
   \label{alg:bas_cmc}
\begin{algorithmic}
   \STATE {\bfseries Input:} current state $s_t$, history vectors $\bm{h}_{s_,a}$ for all $a \in \mathcal{A}$ and $s \in \mathcal{S}$, perception network $\phi$
   \STATE Initialize $bestAction = a_0$
   \STATE Initialize $bestValue = -\infty$
   \FOR{each action $a_i$ in $\mathcal{A}$}
   \STATE $\bm{\alpha} = \phi(s_t, a_i, \bm{h}_{s_t, a_i})$
   \STATE $q(z_{s_t, a_i}) = \text{Dir}(\bm{\alpha})$
   \STATE Current Entropy = $H\Big(q(z_{s_t, a_i})\Big)$
   \STATE Draw reparameterized sample $\tilde{z}_{s_t, a_i}$ from $q(z_{s_t, a_i}|\mathcal{H})$
   \STATE Expected Entropy = 0
   \STATE Expected Future Uncertainty = 0
   \FOR{each state $s_j$ in $\mathcal{S}$}
   \STATE $\bm{h}^{+}_{s_t, a_i} = \bm{h}_{s_t, a_i} + s_j$
   \STATE $q^{+}(z_{s_t, a_i}) = \text{Dir}(\phi(s_t, a_i, \bm{h}^{+}_{s_t, a_i}))$
   \STATE Expected Entropy = Expected Entropy + $\tilde{z}_{s_t, a_i}(j) \times H\Big(q^{+}(z_{s_t, a_i})\Big)$
   \STATE Future Uncertainty = 0
   \FOR{each action $a_k \in \mathcal{A}$}
   \STATE $q(z_{s_j, a_k})$ = $\text{Dir}(\phi(s_j, a_k, \bm{h}_{s_j, a_k}))$
   \STATE Future Uncertainty = Future Uncertainty + $H\Big(q(z_{s_j, a_k})\Big)$
   \ENDFOR
   \STATE Expected Future Uncertainty = Expected Future Uncertainty + $\tilde{z}_{s_t, a_i}(j) \times \text{Future Uncertainty}$
   \ENDFOR
   \STATE $value = (\text{Current Entropy} - \text{Expected Entropy}) + \text{Expected Future Uncertainty}$
   \IF{$value > bestValue$}
   \STATE $bestAction = a_i$
   \ENDIF
   \ENDFOR
\end{algorithmic}
\end{algorithm}
\vfill

\newpage
\vfill
\begin{algorithm}[h]
\caption{Active Sensing in Controllable Markov Chains}
\label{alg:active_sensing_cmc}
\begin{algorithmic}
\STATE Initialize the perception network parameters $\phi_0$ 
\STATE Initialize the history vectors $\bm{h}_{s, a} = \bm{0}$ for all $s \in \mathcal{S}$ and $a \in \mathcal{A}$
\STATE Get initial state $s_0$ from the environment
\FOR{each time step $t \geq 0$}
\STATE Select action $a_t$ with BAS using Algorithm \ref{alg:bas_cmc}
\STATE Execute action $a_t$ and receive the updated state $s_{t+1}$ from the environment
\STATE $\bm{h}_{s_t, a_t} = \bm{h}_{s_t, a_t} + s_{t+1}$
\STATE $q(z_{s_t, a_t}) = \text{Dir}\Big(\phi_t(s_t, a_t, \bm{h}_{s_t, a_t})\Big)$
\STATE Draw reparameterized sample $\tilde{z}_{s_t, a_t}$ from $q(z_{s_t, a_t})$
\STATE $\mathcal{L}_{ELBO} = -\sum_{i=1}^{|\mathcal{S}|} \bm{h}_{s_t, a_t}(i)\log{\tilde{z}_{s_t, a_t}(i)} + D_{KL}\Big(q(z_{s_t, a_t})||\text{Dir}(\bm{1})\Big)$
\STATE Update perception network parameters with gradient descent: $\phi_{t+1} = \phi_{t} + \mu \nabla_{\phi_{t}}\mathcal{L}_{ELBO} $
\ENDFOR
\end{algorithmic}
\end{algorithm}
\vfill

\newpage
\section{Derivation of the ELBO for the perception model}
\label{appendix:av_deriv}
The generative graphical model in Figure \ref{fig:av_model} admits the following factorization of the joint likelihood
\begin{align}
    p(x_{1:T}, z_{1:T}, s|l_{0:T-1}) =&\; p(x_{1:T}|z_{1:T})p(z_{1:T}|l_{0:T-1}, s)p(s)\\
    =&\; p(s)\prod_{t=1}^{T} p(x_{t}|z_{t})\prod_{t=1}^{T} p(z_t|l_{t-1}, s)
\end{align}
Similarly, the joint posterior factorizes as follows
\begin{align}
    q(z_{1:T}, s|x_{1:T}, l_{0:T-1}) =&\; q_1(z_{1:T}|x_{1:T}, l_{0:T-1})q_2(s|z_{1:T}, x_{1:T}, l_{0:T-1})\\
    =&\; q_1(z_{1:T}|x_{1:T}, l_{0:T-1})q_2(s|z_{1:T}, l_{0:T-1})\label{eqn:joint_post_fact}\\
    =&\;q_2(s|z_{1:T}, l_{0:T-1}) \prod_{t=1}^{T}q_1(z_{t}|x_{t}, l_{t-1}), 
\end{align}
We can, therefore, express the log likelihood and posterior probabilities as 
\begin{align}
    \log{p(x_{1:T}, z_{1:T}, s|l_{0:T-1})} 
    =&\; \sum_{t=1}^{T} \log{p(x_{t}|z_{t})} + \sum_{t=1}^{T} \log{p(z_t|l_{t-1}, s)} + \log{p(s)}\label{eqn:log_joint_lik}\\
    \log{q(z_{1:T}, s|x_{1:T}, l_{0:T-1})}
    =&\;  \log{q_2(s|z_{1:T}, l_{0:T-1})} + \sum_{t=1}^{T}\log{q_1(z_{t}|x_{t}, l_{t-1})}\label{eqn:log_joint_post}
\end{align}
Using the log joint likelihood in Equation \ref{eqn:log_joint_lik} and the log posterior in Equation \ref{eqn:log_joint_post}, we can obtain the ELBO on the log marginal likelihood as follows 
\begin{align}
    \log{p(x_{1:T}|l_{0:T-1})} =&\;\log{\mathbb{E}_q\Big[\dfrac{p(x_{1:T}, z_{1:T}, s|l_{0:T-1})}{q(s, z_{1:T}|x_{1:T}, l_{0:T-1})}}\Big]\\
    \geq\;&
    \mathbb{E}_q \Big[\log{p(x_{1:T}, z_{1:T}, s|l_{0:T-1})} - \log{q(s, z_{1:T}|x_{1:T}, l_{0:T-1})} \Big]\label{eqn:jensen}\\
    \begin{split}
    =&\; \mathbb{E}_q \Big[ \sum_{t=1}^{T} \log{p(x_{t}|z_{t})} + \sum_{t=1}^{T} \log{p(z_t|l_{t-1}, s)} + \log{p(s)} 
    \\&\quad\quad- \sum_{t=1}^{T}\log{q_1(z_{t}|x_{t}, l_{t-1})} - \log{q_2(s|z_{1:T}, l_{0:T-1})}\Big]
    \end{split}\\
    \begin{split}
    =&\;  \sum_{t=1}^{T} \mathbb{E}_q \Big[\log{p(x_{t}|z_{t})}\Big] + \sum_{t=1}^{T} \mathbb{E}_q \Big[\log{\dfrac{p(z_t|l_{t-1}, s)}{q_1(z_t|x_t, l_{t-1})}}\Big] \\&\quad\quad + \mathbb{E}_q \Big[ \log{\dfrac{p(s)}{q_2(s|z_{1:T}, l_{0:T-1})}}\Big], \label{eqn:final_elbo}        
    \end{split}
\end{align}
where Equation \ref{eqn:jensen} follows from Jensen's inequality.

\newpage
\section{Survey of related work}
\label{appendix:related_work}
\paragraph{Intrinsic motivation} Our approach can be regarded as an intrinsically-motivated exploration strategy \cite{barto_im}. In intrinsically-motivated exploration, an agent learns exploratory behavior in the absence of any extrinsic reward signals. Instead of extrinsic reward, exploration is guided by intrinsic value, which in our case is based on the expected uncertainty reduction associated with an action. The uncertainty is measured with respect to the agent's perception model, which is learned in a completely unsupervised manner. Other types of intrinsic signals have been used for autonomous exploration, such as prediction error \cite{schmidhuber1991, pathak_ICM}, space coverage \cite{hazan_space_coverage, amin_space_coverage}, and visitation count \cite{menard_vis_count}. Intrinsic motivation strategies offer the advantage of representations that generalize to different tasks in the same environment since there is no dependence on a specific reward function. The closest family of intrinsic motivation approaches to ours are information-theoretic approaches, discussed below. 

\paragraph{Information-theoretic exploration in reinforcement learning} Information gain has been used to promote autonomous exploration in multiple approaches such as \cite{storck1995, sun2011, still2012}. However, these approaches rely on state-action enumeration to compute information gain, which limits their applicability to settings with discrete state and action spaces. In contrast, our framework is general and can be applied to both discrete and continuous settings. In the discrete setting, our work is most related to \cite{little_sommer}; we test our model in a maze navigation task similar to the one used there. The main difference between their approach and ours is that we do not assume explicit knowledge about the true generative model of the environment. Instead, the perception component of our architecture learns a generative model through collected experiences in an end-to-end manner. In the continuous setting, our work is most similar to \cite{vime} and \cite{mohamed_variational_infomax} in deep reinforcement learning (RL). Our approach is different from those two approaches in that it can be applied in model-based settings since the perception component of our model explicitly learns the transition dynamics of the environment, enabling the generation of \textit{imagined} trajectories that can be used for model-based planning and training. In contrast, those two approaches rely on model-free methods by modifying the reward function to include an information gain component. 

\paragraph{Active vision and visual attention in machine learning} We apply our model to the task of active vision. Here, our work is related to the Recurrent Attention Model (RAM) by \cite{mnih_ram} and the DRAW model by \cite{gregor_draw}, but differs from those models in four key aspects. First, the perception and action components of our model are trained in a completely unsupervised, task-independent manner. During the classification task, only one feedforward decision network (separate from the main model) is trained with the classification loss. The learned representations can then be used for arbitrary tasks: to illustrate, we use these representations as input to the decision network to achieve high performance on a downstream image classification task. Second, despite the sequential nature of this task, our model solves it using end-to-end feedforward networks, greatly reducing the amount of computation compared to the recurrent architectures used in \cite{mnih_ram} and \cite{gregor_draw}. Third, in contrast to \cite{gregor_draw}, our model does not assume access to the full image in the training loss function, which is consistent with the assumption of bandlimited sensing. Finally, our model makes explicit links to ideas in neuroscience that enable the testing of functional hypotheses in a modern machine learning setting. 

Our active vision approach is also somewhat related to the framework of Attend, Infer, Repeat (AIR) \citep{air, sqair}. However, there are some key differences in the underlying modeling assumptions. For example, the latent variables inferred by AIR for a given image (or scene) are assumed to correspond to the attributes of entire objects that decompose the scene, whereas our model makes no restrictions on what the lower-level latent variables represent. Further, for biological plausibility, our approach imposes the restriction that only patches of the image corresponding to the model's bandlimited sensor are used during inference and training. Since AIR is mainly concerned with decomposing scenes into constituent objects, it makes no such restriction and so the underlying approach involves a different inference and training process. 

\paragraph{Active inference and the free energy principle} In general, the theoretical formulation of our approach is most similar to the active inference formulation in neuroscience \cite{friston_fep, active_inference_process_theory}. However, there are two differences. First, action selection in active inference relies on minimizing a generalized Expected Free Energy (EFE), whereas we use a more specific uncertainty reduction objective geared towards exploratory behavior. Second, current implementations of active inference use enumerated trajectories to minimize EFE, which limits their applicability to discrete state and action spaces. In contrast, our approach combines perception and action into an integrated and scalable neural network model that is easily applicable to diverse tasks.
\paragraph{Predictive coding in machine learning} There is a large body of work adapting the theory of predictive coding to machine learning problems, ranging from computer vision \cite{pred_net, deepPCN, ororbia_conv}, gradient-based optimization \cite{salvatori_predBackProp, millidge_backprop}, lifelong learning \cite{lifelong_learning_ororbia}, and temporal learning \cite{ororbia_temporal_learning}. However, these models apply the theory in the context of passive perception. Although some recent work combines predictive coding models with action \cite{ororbia_angc}, they do not focus on autonomous exploration. Rao and colleagues have recently introduced the framework of active predictive coding \cite{rao_apc_1, rao_apc_2, rao_apc_3, fisher_rnp} but their generative models focus on generating transition and policy functions using hypernetworks, and their policy training approach utilizes a supervised reinforcement learning algorithm without particular focus on autonomous exploration.

\newpage
\section{Experiment settings and hyperparameters for controllable Markov chains}
\label{appendix:cmc_hp}
\subsection{Dense World}
In the dense world experiments, we let the models explore an environment with 10 states and 4 actions for a total of 2000 timesteps. The transition distributions in the environment were independently drawn from a Dirichlet distribution with parameter $\bm{\alpha} = 1$. That is, for a given state $s$ and action $a$, the transition distribution $p(:|s, a)$ is given by 
\begin{align}
    p(:|s, a) \sim \text{Dir(1)} := \dfrac{\Gamma(N)}{\Gamma(1)^{N}}\\
    \Gamma(x) := \int_{0}^{\infty} t^{x - 1}e^{-t}\;dt
\end{align}
For both the BAS and the random agents, the perception network was a two-layer feedforward network with 16 hidden units each followed by a softplus nonlinearity. The learning rate for both agents was fixed at 0.001.  

\subsection{Mazes}
In the maze experiments, the agents explored a maze for a total of 3000 time steps. We did experiments with different maze sizes ($6\times6, \; 8\times8$, and $12\times 12$) and observed qualitatively the same results. We used the same network architecture and learning rates for all agents as those used in the Dense World experiments. 

\newpage
\section{Experiment settings and hyperparameters for active vision}
\label{appendix:av_hp}

\subsection{MNIST classification}
\label{sec:mnist_hypers}
We trained our model on the regular MNIST dataset, the translated MNIST dataset (described in the main text), and the fashion MNIST dataset. First, the perception model was pre-trained in a completely unsupervised manner with randomly selected fixation locations. Afterwards, we continued to train the perception model with the unsupervised loss while actions were selected using our BAS strategy. At the same time, a separate decision network was trained to take as input the inferred state $s$ at the end of trial output a class label at the end of the trial. Gradients from classification loss were only used to update the parameters of the decision network. For all experiments, the encoder and decoder networks of both the the lower-level and the higher-level VAEs were feedforward networks with two layers, each with 256 hidden units followed by rectified linear unit (ReLU) activation functions. The action network was a two-layer feedforward network with 64 and 32 hidden units, respectively. When perception states were used for decision making, the decision network was a two-layer feedforward network with 256 hidden units each. When an RNN was used to integrate past observations for decision-making, the hidden size of the RNN decision network was chosen to be the same as the dimensionality of the abstract state $s$. Table \ref{table:mnist_settings} lists the hyperparameters used for each type of experiment. Hyperparameters were adjust ad hoc based on the resulting accuracy obtained on a separate validation set. In these experiments, we also use a regularization hyperparameter $\beta$ as a scalar multiplying the KL term in the ELBO objective for the perception model \cite{beta_vaes}.  

\begin{table}[h]
    \caption{Settings for Centered, Translated, and Fashion MNIST Experiments}
    \vskip 0.1in
    \centering
    \label{table:mnist_settings}
    \begin{tabular}{lllll}
    \toprule
     Hyper-parameter & Centered MNIST & Translated MNIST & Fashion MNIST \\
    \midrule
    $\#$ pre-training episodes & 0 & 10 & 10\\
     $\#$ fixations (n)    & 3 & 4 & 5\\
     Patch dim (d)     &  8 & 12 & 6\\
     $\#$ foveated patches ($N_{fov}$) & 1 & 3 & 1\\
     Foveation scale   & --- & 2 & ---\\
     z dim             & 32 & 64 & 64\\
     s dim             & 64 & 128 & 128\\
     $\sigma_{action}$ & 0.15 & 0.15 & 0.05\\
     Action network lr & 0.001 & 0.001 & 0.001\\
     Perception model lr & 0.001 & 0.001 & 0.001\\
     Decision network lr & 0.001 & 0.001 & 0.001\\
     $\beta$             & 0.1 & 0.1 & 1.1\\
     Batch size      & 64 & 64 & 64\\
    \bottomrule
    \end{tabular}
\end{table}

\subsection{Grayscale CIFAR-10}
We tested our perception model on grayscale CIFAR-10 images to see if it can capture the overall structure and statistics of natural images. Table \ref{table:cifar_settings} lists the hyperparameters used for these experiments.\begin{table}[h]
    \caption{Settings for Grayscale CIFAR-10 Experiments}
    \vskip 0.1in
    \centering
    \label{table:cifar_settings}
    \begin{tabular}{ll}
    \toprule
     Hyper-parameter & Setting \\
    \midrule
     $\#$ fixations (n)    & 6\\
     Patch dim (d)     &  12\\
     $\#$ foveated patches ($N_{fov}$) & 1\\
     z dim             & 32\\
     s dim             & 64\\
     Perception model lr & 0.001\\
     $\beta$             & 0.01\\
     Batch size      & 64\\
    \bottomrule
    \end{tabular}
\end{table}

\subsection{Learning speed and data efficiency comparisons}
All tests reported in Section \ref{sec:data_eff} were performed on the translated MNIST dataset. Our approach (BAS + FF), described in the main text, was compared to four baselines: the Recurrent Attention Model (RAM) \citep{mnih_ram}, VIME \citep{vime}, Plan2Explore \citep{p2e}, and a feedforward (FF) neural network receiving full images as input (Full Images + FF). In all these cases, the decision network (the network that outputs class labels) consisted of two hidden layers each with 128 hidden units followed by ReLU activation functions. 

BAS + FF, VIME, and Plan2Explore were all pre-trained unsupervised for 10 epochs with a random action selection strategy. For BAS + FF, the architectures of the perception model and the action network were the same as those described in Section \ref{sec:mnist_hypers}. For RAM, the hidden size of the RNN was chosen to match the dimensionality of the abstract representation $s$ in our perception model, which was 128. In all models (except Full Images + FF), the location was drawn from a two-component Gaussian (with a pre-determined fixed variance) parameterized by the output of the location network. Hyperparameters for all models were adjusted ad hoc to optimize performance on a validation set (separate from the MNIST test set). 

All hyperparameters are listed in Tables \ref{table:data_eff_global_hp}, \ref{table:data_eff_bas_hp}, \ref{table:data_eff_ram_hp}, \ref{table:data_eff_vime_hp}, and \ref{table:data_eff_p2e_hp}. In the Full Images + FF case, the only hyperparameters adjusted were the batch size and the learning rate which were fixed at 64 and 0.001, respectively.

\begin{table}[h]
\caption{Global Hyperparameters}
\centering
\vskip 0.1in
\begin{tabular}{ll}
\toprule
 Hyper-parameter & Value \\
\midrule
 $\#$ fixations (n)    & 3 \\
 Patch dim (d)     &  12 \\
 $\#$ foveated patches ($N_{fov}$) & 3 \\
 Foveation scale & 2 \\
 Batch size      & 64 \\
\bottomrule
\end{tabular}
\vskip 0.2in
\label{table:data_eff_global_hp}
\end{table}

\begin{table}[h]
\caption{Hyperparameters for BAS + FF}
\centering
\vskip 0.1in
\begin{tabular}{ll}
\toprule
 Hyper-parameter & Value \\
\midrule
 $z$ dim & 64\\
 $s$ dim & 128\\
 $\sigma_{action}$ & 0.15\\
 Action network hidden layers & [64] \\
 Action network lr & 0.001\\
 Perception model lr & 0.001\\
 Decision network lr & 0.001\\
 $\beta$ & 0.1\\     
\bottomrule
\end{tabular}
\vskip 0.2in
\label{table:data_eff_bas_hp}
\end{table}

\begin{table}[h]
\caption{Hyperparameters for RAM}
\centering
\vskip 0.1in
\begin{tabular}{ll}
\toprule
 Hyper-parameter & Value \\
\midrule
 $h_g$ & 64\\
 $h_l$ & 64\\
 RNN hidden size & 128\\
 $\sigma_{action}$ & 0.05\\
 Action network hidden layers & [64] \\
 Action network lr & 0.001\\
 Decision network lr & 0.001 \\
 Core and glimpse networks lr & 0.001\\
\bottomrule
\end{tabular}
\vskip 0.2in
\label{table:data_eff_ram_hp}
\end{table}

\begin{table}[h]
\caption{Hyperparameters for VIME}
\centering
\vskip 0.1in
\begin{tabular}{ll}
\toprule
 Hyper-parameter & Value \\
\midrule

Action network hidden layers & [256] \\
$\sigma_{action}$ & 0.05 \\
BNN hidden size & 256 \\ 
BNN prior std & 0.5 \\
BNN likelihood std & 5.0 \\
BNN lr & 0.0001 \\
Action network lr & 0.001 \\
Decision network lr & 0.001\\

\bottomrule
\end{tabular}
\vskip 0.2in
\label{table:data_eff_vime_hp}
\end{table}

\begin{table}[h]
\caption{Hyperparameters for Plan2Explore}
\centering
\vskip 0.1in
\begin{tabular}{ll}
\toprule
 Hyper-parameter & Value \\
\midrule

Number of ensemble models & 10 \\
One step model layers & [256, 256] \\
Horizon length & 15 \\
Encoder layers & [256, 256] \\ 
Decoder layers & [256, 256] \\ 
Action network hidden layers & [256, 256] \\
Critic network hidden layers & [256, 256]\\ 
Reward network layers & [32, 32]\\
$\sigma_{action}$ & 0.01 \\
Action network lr & 0.001 \\
Critic network lr & 0.001 \\
One step model lr & 0.001 \\
Decision network lr & 0.001\\

\bottomrule
\end{tabular}
\vskip 0.2in
\label{table:data_eff_p2e_hp}
\end{table}

\clearpage
\section{Supplementary figures}
\vskip 1 in
\begin{figure}[h]
    \centering
    \includegraphics[width=\textwidth]{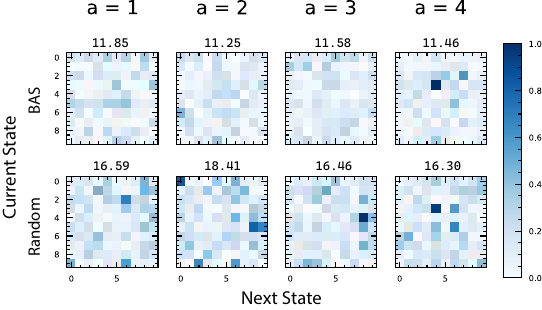}
    \caption{Differences between learned and true transition distributions for our active explorer (BAS, top row) and a random explorer (bottom row) in the Dense Worlds environment. Each image represents the normalized absolute differences between the learned and true transition distributions for a given action (a = 1, 2, 3, 4). Numbers above each image represent the sum of all values in the corresponding matrix. In each case, agents were allowed to explore the environment for 2000 steps.}
    \label{fig:prob_diffs}
\end{figure}
\vfill

\begin{figure}[h]
    \centering
    \begin{subfigure}{0.45\textwidth}
        \includegraphics[width=\textwidth]{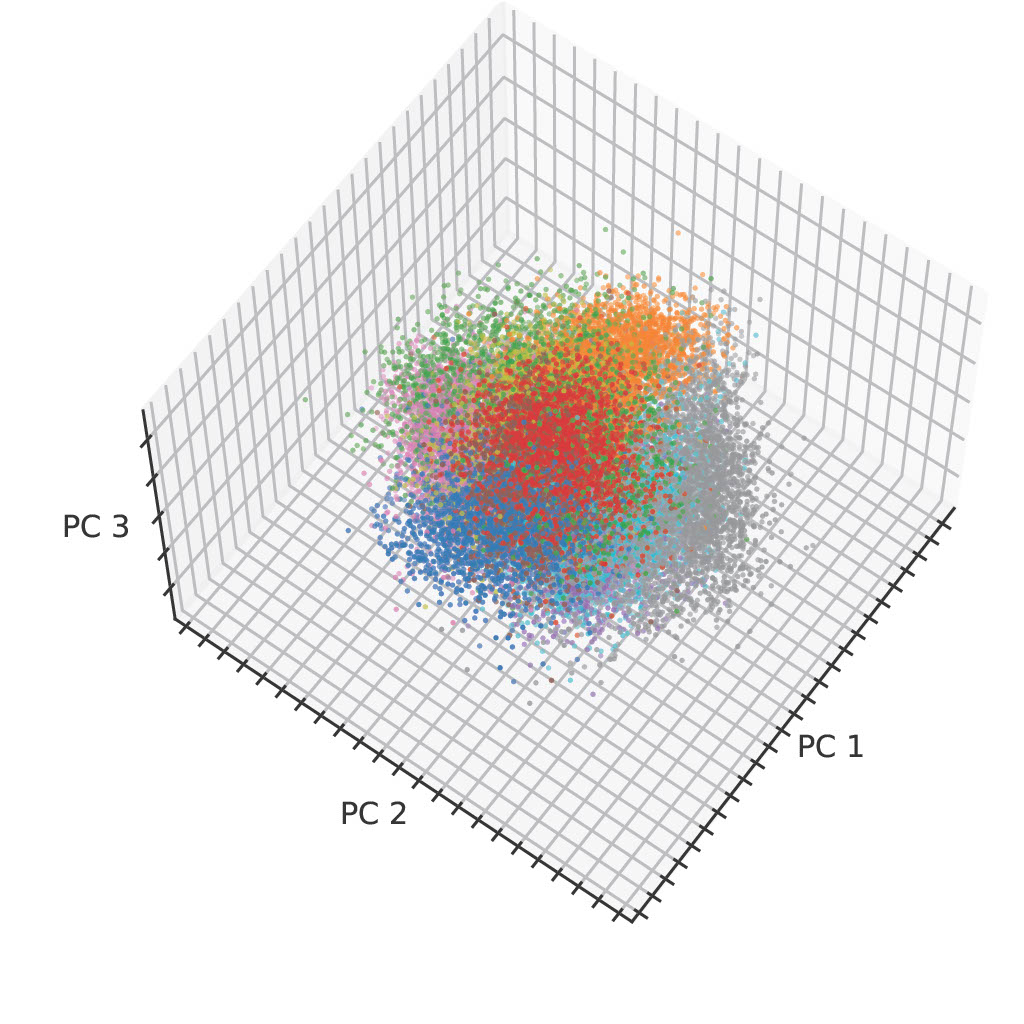}
        \caption{}
        \label{fig:bas_pca}
    \end{subfigure}
    \;
    \begin{subfigure}{0.45\textwidth}
        \includegraphics[width=\textwidth]{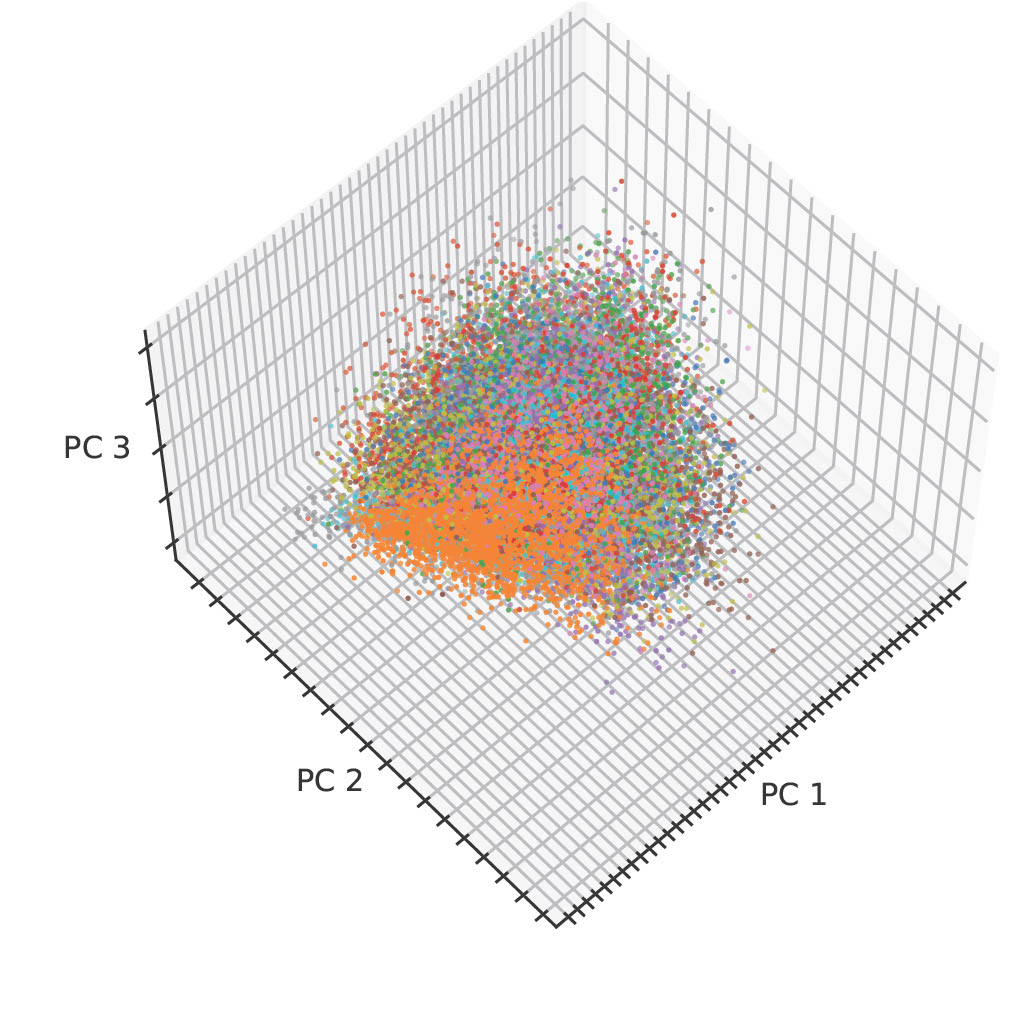}
        \caption{}
        \label{fig:random_pca}
    \end{subfigure}
    \caption{PCA projections of the latent representations learned by (a) BAS-trained perception model, and (b) the randomly-trained perception model. Each point in the PC space correspond the projection of the inferred state $s$ for a given input image. Points are colored based on the class of their corresponding input images.}
    \label{fig:perception_pca}
\end{figure}

\begin{figure}[h]
    \centering
    \begin{subfigure}{0.45\textwidth}
        \includegraphics[width=\textwidth]{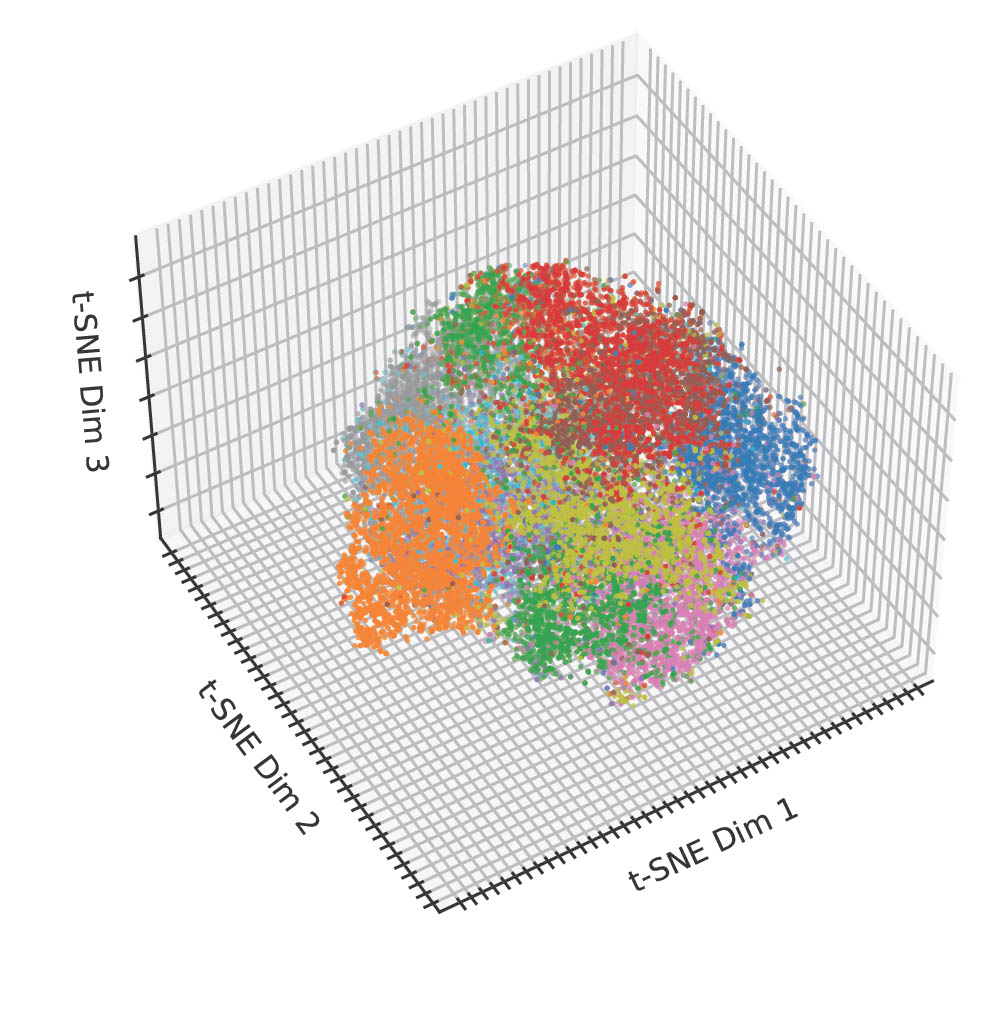}
        \caption{}
        \label{fig:bas_tsne}
    \end{subfigure}
    \;
    \begin{subfigure}{0.47\textwidth}
        \includegraphics[width=\textwidth]{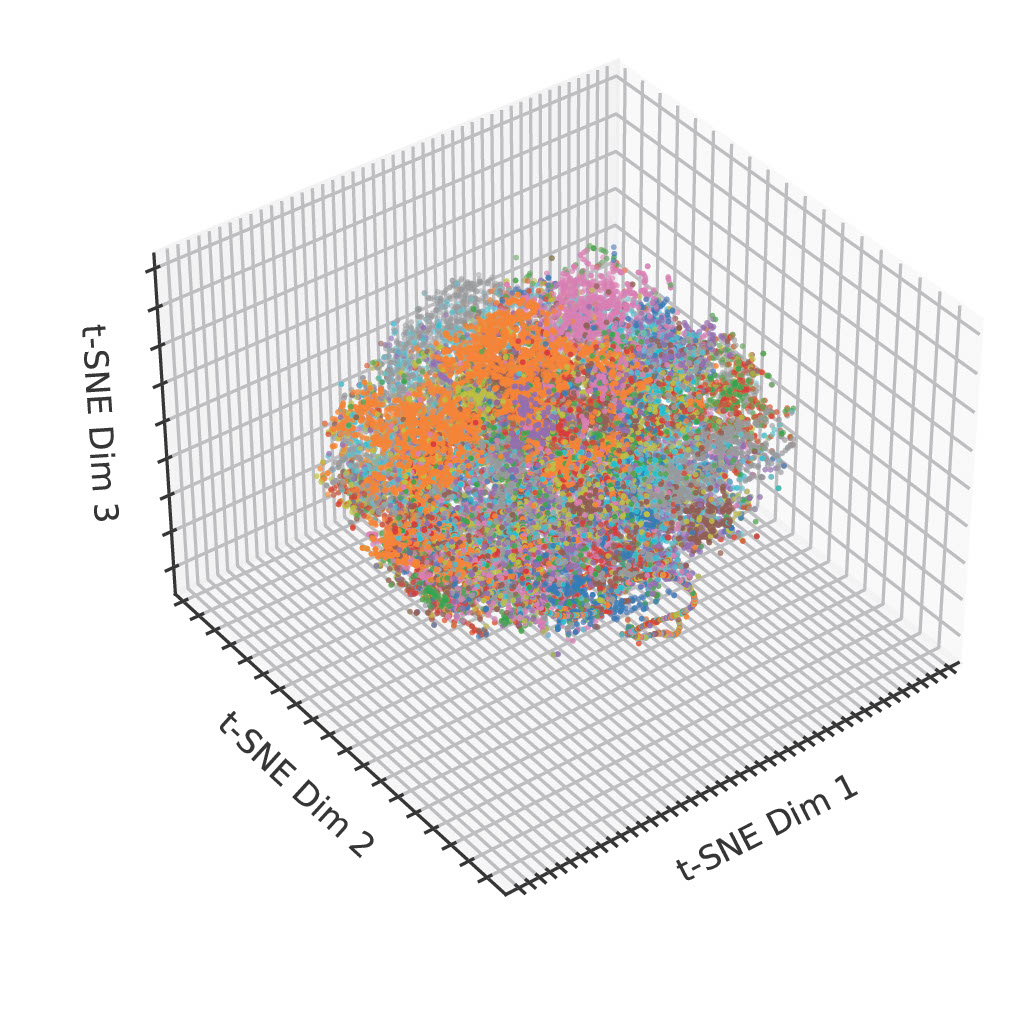}
        \caption{}
        \label{fig:random_tsne}
    \end{subfigure}
    \caption{Same as in Figure \ref{fig:perception_pca} but using t-SNE \cite{tsne} to visualize projections. (a) t-SNE projections of the BAS-trained perception model. (b) t-SNE projections for the randomly-trained perception model.}
    \label{fig:perception_tsne}
\end{figure}
\vfill

\newpage
\vfill
\begin{figure}[h]
    \centering
    \vskip 0.8in
    \hskip 0.8in
    \includegraphics[width=0.8\textwidth]{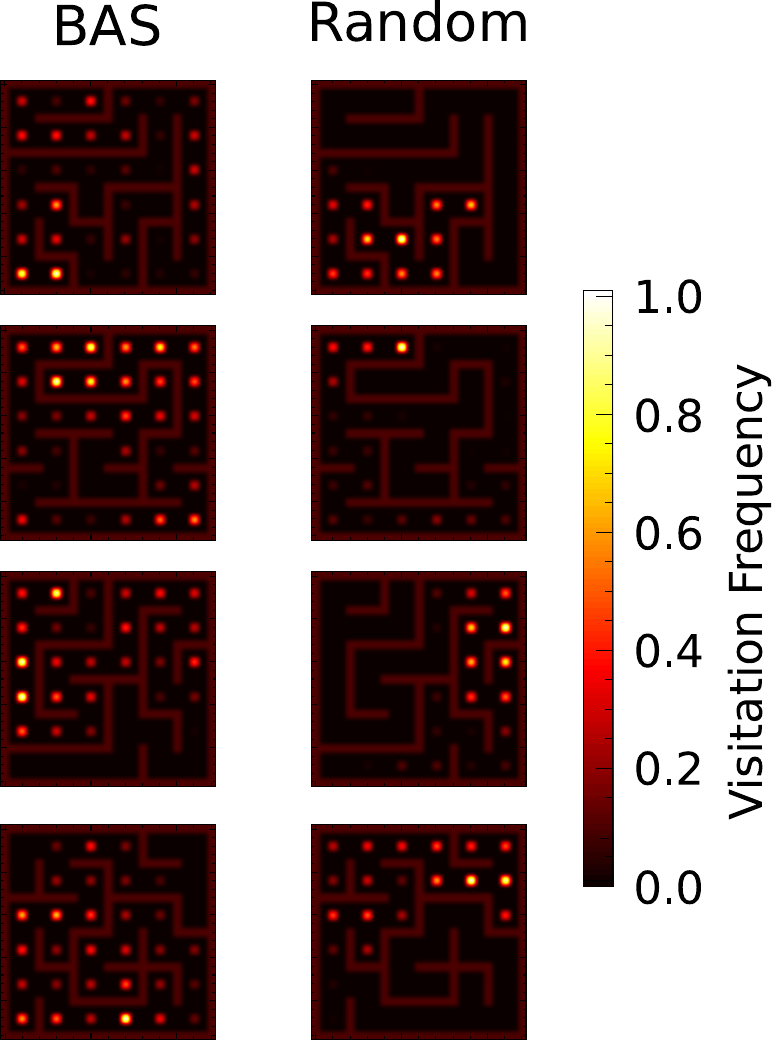}
    \vskip 0.2in
    \caption{More examples of visitation frequency heatmaps for our BAS explorer model versus a random exploration strategy in $6\times6$ mazes.}
    \label{fig:more_maze_heatmaps_6x6}
\end{figure}
\vfill

\newpage
\vfill
\begin{figure}[h]
    \centering
    \begin{subfigure}{0.5\textwidth}
        \hskip -0.15in
        \includegraphics[width=\textwidth]{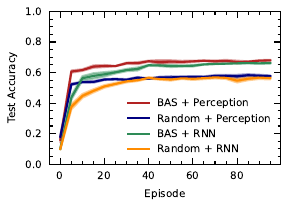}
        \caption{}
        \label{fig:fmnist_perf}
    \end{subfigure}
    \\
    \vskip 0.1in
    \begin{subfigure}{0.55\textwidth}
        \includegraphics[width=\textwidth]{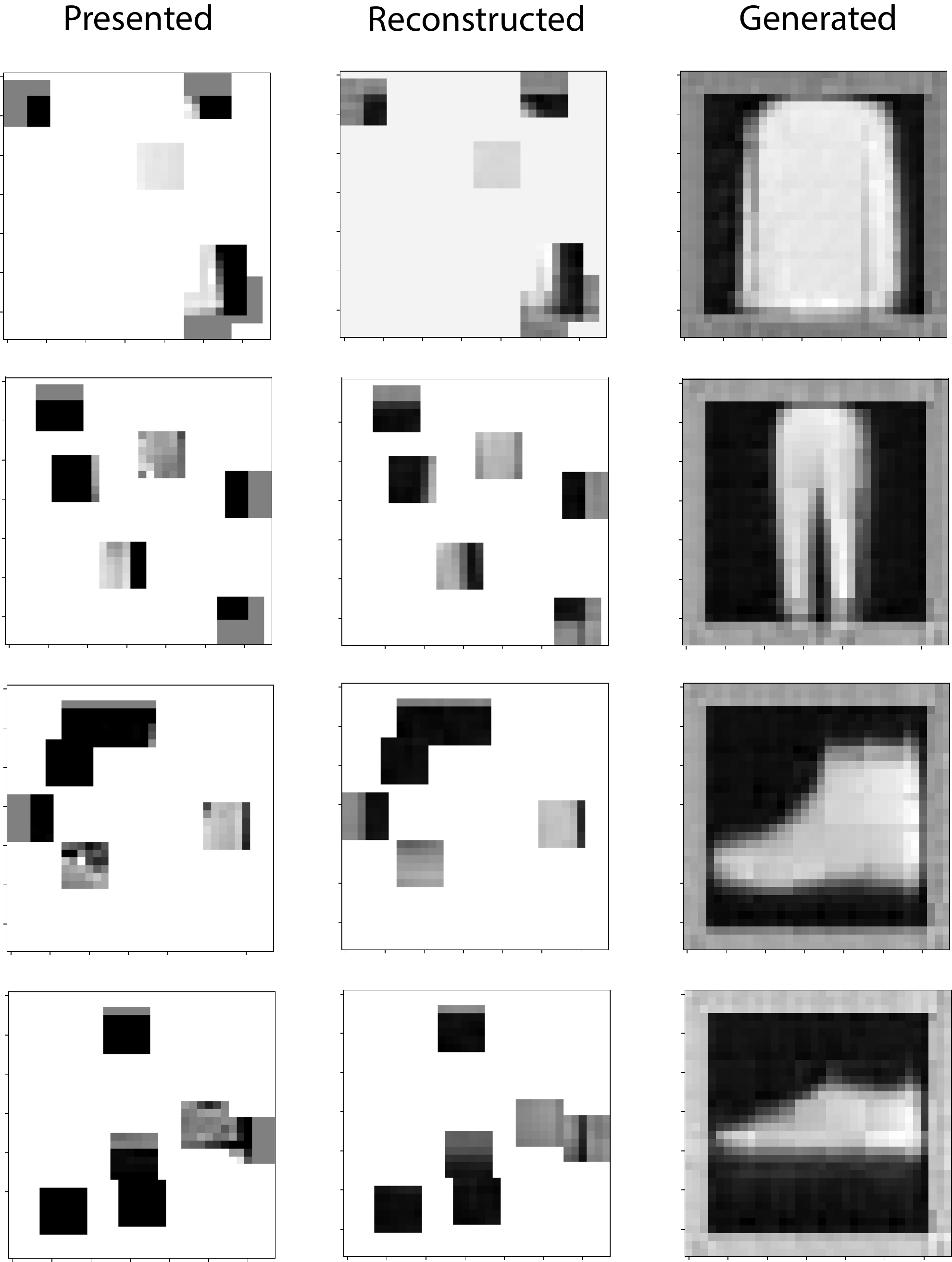}
        \caption{}
        \label{fig:fmnist_samples}
    \end{subfigure}
    \caption{Results on Fashion MNIST. (a) Classification performance on the fashion MNIST dataset during the active vision task. (b) Examples demonstrating the generative ability of the perception model. The original patches presented are shown on the left and their reconstructions are shown in the middle. Right shows images generated by first generating small patches at various locations (not seen during presentation) and combining them to form the final image. These results show the perception model is able to capture the spatial relationships associated with elements in the dataset.}
    \label{fig:fmnist_results}
\end{figure}
\vfill
\clearpage

\newpage
\vfill
\begin{figure}[h]
    \centering
    \includegraphics[width=0.9\textwidth]{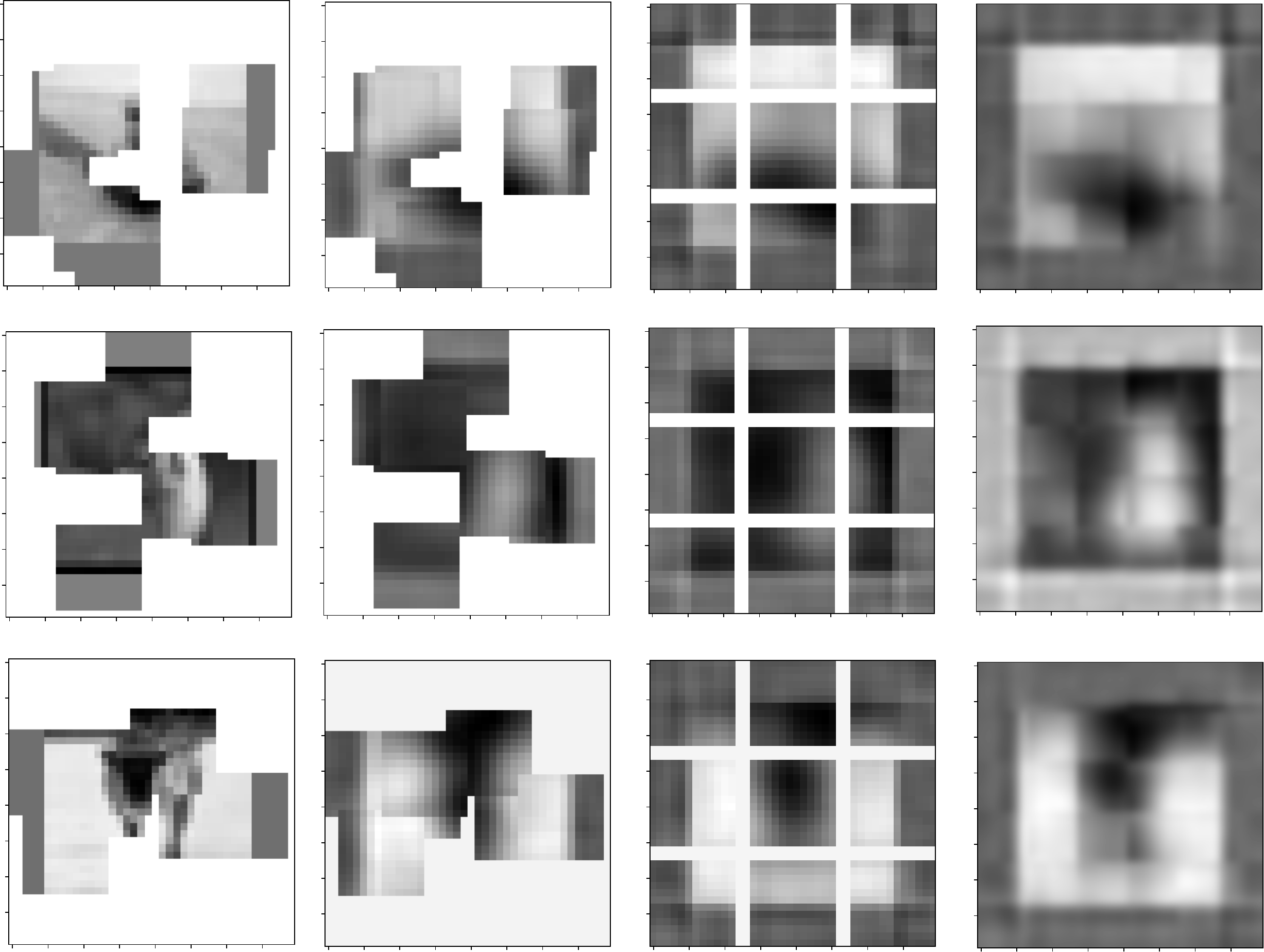}
    \caption{Results on grayscale CIFAR-10. Original presented patches are shown on the left and their reconstructions are shown on the second column. Third and fourth column show images generated by the perception model combining smaller patches at different locations. The last column has more patches at continguous locations followed by bicubic smoothing for illustration. These results show that, despite the simplicity of our model's architecture, it is still able to capture the overall structure and statistics in natural images.}
    \label{fig:cifar_samples}
\end{figure}
\vfill

\clearpage

%%%%%%%%%%%%%%%%%%%%%%%%%%%%%%%%%%%%%%%%%%%%%%%%%%%%%%%%%%%%%%%
\bibliographystyle{apalike}
\bibliography{paper}

\end{document}